\documentclass[a4paper,journal,compsoc]{IEEEtran}

\usepackage{microtype}
\usepackage{graphicx}
\usepackage{subfigure}
\usepackage{booktabs} 
\usepackage{csquotes}
\usepackage{cite}
\usepackage[T1]{fontenc}
\usepackage{hyperref}
\usepackage[normalem]{ulem}
\usepackage{xcolor}
\usepackage{amsmath,amssymb,amsthm}
\newtheorem{thm}{Theorem}
\newtheorem{lem}{Lemma}
\theoremstyle{definition}

\usepackage{footmisc}
\usepackage{verbatim}

\usepackage{tikz}
\usetikzlibrary{shapes.misc}
\usepackage{pgfplots}

\newcommand{\asymcloud}[2][.1]{%
	\begin{scope}[#2]
		\pgftransformscale{#1}%
		\pgfpathmoveto{\pgfpoint{261 pt}{115 pt}} 
		\pgfpathcurveto{\pgfqpoint{70 pt}{107 pt}}
		{\pgfqpoint{137 pt}{291 pt}}
		{\pgfqpoint{260 pt}{273 pt}} 
		\pgfpathcurveto{\pgfqpoint{78 pt}{382 pt}}
		{\pgfqpoint{381 pt}{445 pt}}
		{\pgfqpoint{412 pt}{410 pt}}
		\pgfpathcurveto{\pgfqpoint{577 pt}{587 pt}}
		{\pgfqpoint{698 pt}{488 pt}}
		{\pgfqpoint{685 pt}{366 pt}}
		\pgfpathcurveto{\pgfqpoint{840 pt}{192 pt}}
		{\pgfqpoint{610 pt}{157 pt}}
		{\pgfqpoint{610 pt}{157 pt}}
		\pgfpathcurveto{\pgfqpoint{531 pt}{39 pt}}
		{\pgfqpoint{258 pt}{51 pt}}
		{\pgfqpoint{261 pt}{115 pt}}
		\pgfusepath{fill,stroke}         
\end{scope}}

\newcommand{\asymcloudd}[2][.1]{%
	\begin{scope}[#2]
		\pgftransformscale{#1}%
		\pgfpathmoveto{\pgfpoint{261 pt}{115 pt}} 
		\pgfpathcurveto{\pgfqpoint{70 pt}{107 pt}}
		{\pgfqpoint{137 pt}{291 pt}}
		{\pgfqpoint{260 pt}{273 pt}} 
		\pgfpathcurveto{\pgfqpoint{78 pt}{382 pt}}
		{\pgfqpoint{381 pt}{445 pt}}
		{\pgfqpoint{412 pt}{410 pt}}
		\pgfpathcurveto{\pgfqpoint{577 pt}{587 pt}}
		{\pgfqpoint{698 pt}{488 pt}}
		{\pgfqpoint{685 pt}{366 pt}}
		\pgfpathcurveto{\pgfqpoint{400 pt}{192 pt}}
		{\pgfqpoint{500 pt}{57 pt}}
		{\pgfqpoint{500 pt}{57 pt}}
		\pgfpathcurveto{\pgfqpoint{531 pt}{39 pt}}
		{\pgfqpoint{258 pt}{51 pt}}
		{\pgfqpoint{261 pt}{115 pt}}
		\pgfusepath{fill,stroke}         
\end{scope}}

\newcommand{\asymclouddd}[2][.1]{%
	\begin{scope}[#2]
		\pgftransformscale{#1}%
		\pgfpathmoveto{\pgfpoint{261 pt}{115 pt}} 
		\pgfpathcurveto{\pgfqpoint{70 pt}{107 pt}}
		{\pgfqpoint{137 pt}{291 pt}}
		{\pgfqpoint{260 pt}{273 pt}} 
		\pgfpathcurveto{\pgfqpoint{100 pt}{200 pt}}
		{\pgfqpoint{445 pt}{445 pt}}
		{\pgfqpoint{410 pt}{410 pt}}
		\pgfpathcurveto{\pgfqpoint{577 pt}{587 pt}}
		{\pgfqpoint{698 pt}{488 pt}}
		{\pgfqpoint{685 pt}{366 pt}}
		\pgfpathcurveto{\pgfqpoint{425 pt}{192 pt}}
		{\pgfqpoint{550pt}{57 pt}}
		{\pgfqpoint{605 pt}{120 pt}}
		\pgfpathcurveto{\pgfqpoint{300 pt}{60 pt}}
		{\pgfqpoint{258 pt}{51 pt}}
		{\pgfqpoint{261 pt}{115 pt}}
		\pgfusepath{fill,stroke}         
\end{scope}}

\newcommand{\asymcloudddd}[2][.1]{%
	\begin{scope}[#2]
		\pgftransformscale{#1}%
		\pgfpathmoveto{\pgfpoint{261 pt}{115 pt}} 
		\pgfpathcurveto{\pgfqpoint{78 pt}{382 pt}}
		{\pgfqpoint{381 pt}{445 pt}}
		{\pgfqpoint{412 pt}{410 pt}}
		\pgfpathcurveto{\pgfqpoint{577 pt}{587 pt}}
		{\pgfqpoint{698 pt}{488 pt}}
		{\pgfqpoint{685 pt}{366 pt}}
		\pgfpathcurveto{\pgfqpoint{531 pt}{39 pt}}
		{\pgfqpoint{258 pt}{51 pt}}
		{\pgfqpoint{261 pt}{115 pt}}
		\pgfusepath{fill,stroke}         
\end{scope}}

\begin{document}

\newcommand{\ent}[1]{H(#1)}
\newcommand{\diffent}[1]{h(#1)}
\newcommand{\mutinf}[1]{I(#1)}
\newcommand{\kl}[2]{D\left(#1\|#2\right)}
\newcommand{\cent}[2]{C\left(#1\|#2\right)}
\newcommand{\expop}[2]{E_{ #1}\left[#2\right]}
\newcommand{\infodim}[1]{d(#1)}
\newcommand{\barmutinfc}[1]{\bar{I}_C(#1)}
\newcommand{\barmutinfp}[1]{\bar{I}_P(#1)}
\newcommand{\estmutinf}[1]{\hat{I}(#1)}
\newcommand{\Prob}[1]{\mathbb{P}(#1)}
\newcommand{\qRVm}[1]{[#1]_m}
\newcommand{\pmf}[1]{p_{#1}}
\newcommand{\enttwo}[1]{H^2(#1)}
\newcommand{\cordim}[1]{d^2(#1)}
\newcommand{\cube}[2]{\mathcal{C}(#1,#2)}
\newcommand{\dom}[1]{\mathcal{#1}}
\newcommand{\reals}{\mathbb{R}}
\newcommand{\outputguess}{\hat{Y}}
\newcommand{\lastlayer}{\tilde{Y}}

\title{Learning Representations for Neural Network-Based Classification Using the Information Bottleneck Principle} 

\author{Rana Ali Amjad,~\IEEEmembership{Student Member,~IEEE,} %
and Bernhard C. Geiger,~\IEEEmembership{Senior Member,~IEEE}%
\IEEEcompsocitemizethanks{\IEEEcompsocthanksitem Rana Ali Amjad is with the Institute for Communications Engineering, Technical University of Munich, Germany. Email: ranaali.amjad@tum.de\IEEEcompsocthanksitem Bernhard C. Geiger was with the Signal Processing and Speech Communication Laboratory, Graz University of Technology, Austria and is now with Know-Center GmbH, Graz, Austria. Email: geiger@ieee.org}}

\IEEEtitleabstractindextext{
\begin{abstract}
In this theory paper, we investigate training deep neural networks (DNNs) for classification via minimizing the information bottleneck (IB) functional. We show that the resulting optimization problem suffers from two severe issues: First, for deterministic DNNs, either the IB functional is infinite for almost all values of network parameters, making the optimization problem ill-posed, or it is piecewise constant, hence not admitting gradient-based optimization methods. Second, the invariance of the IB functional under bijections prevents it from capturing properties of the learned representation that are desirable for classification, such as robustness and simplicity. We argue that these issues are partly resolved for stochastic DNNs, DNNs that include a (hard or soft) decision rule, or by replacing the IB functional with related, but more well-behaved cost functions. We conclude that recent successes reported about training DNNs using the IB framework must be attributed to such solutions. As a side effect, our results indicate limitations of the IB framework for the analysis of DNNs. We also note that rather than trying to repair the inherent problems in the IB functional, a better approach may be to design regularizers on latent representation enforcing the desired properties directly. 
\end{abstract}
}

\maketitle

\section{Introduction}\label{sec:intro}
Recently, the information bottleneck (IB) framework has been proposed for analyzing and understanding DNNs~\cite{Tishby_DLIB_ITW}. The IB framework admits evaluating the optimality of the learned representation and has been used to make claims regarding properties of stochastic gradient descent optimization and the computational benefit of many hidden layers~\cite{Tishby_BlackBox}. Whether these claims all hold true is the subject of an ongoing debate (cf.~\cite{Anonymous_IBTheory}).

Rather than adding to this debate, the purpose of this paper is to contribute to a different research area sparked by~\cite{Tishby_DLIB_ITW}: Training DNNs for classification by minimizing the IB functional. To be more precise, suppose that $Y$ is a class variable, $X$ are features at the input of the DNN, and $L$ is either a latent representation or the output of the DNN for the input $X$. The IB functional then is~\cite{Tishby_InformationBottleneck}
\begin{equation}\label{eq:IB}
\mutinf{X;L}-\beta\mutinf{Y;L} 
\end{equation}
for some trade-off parameter $\beta>1$. A DNN minimizing this functional thus has a maximally compressed latent representation or output $L$ (because the mutual information $\mutinf{X;L}$ is small) that is informative about the class variable (because $\mutinf{Y;L}$ is large). The IB framework thus introduces a regularization term $\mutinf{X;L}$ that depends on the representation $L$ rather than on the parameters of the DNN. In general data-dependent regularization has the potential to capture properties in the latent representations or the output of the DNN desirable for the specific classification task; namely, robustness against noise and small distortions and simplicity of the representation (see Sec.~\ref{sec:regularization} for details). In this paper, we will investigate if IB functional-based data dependent regularization can do this or not. 

Subsequently, the IB framework has been used to train DNNs for discrete or continuous features~\cite{Kolchinsky_NLIB,Alemi_DVIB,Anonymous_ParametricIB, uncertaintyDVIB, vdb}. These works report remarkable performance in classification tasks (see also Sec.~\ref{sec:related}), but only after slightly departing from the IB framework by claiming that~\eqref{eq:IB} is hard to compute. As a remedy, they replace mutual information terms with bounds in order to obtain cost functions that can be computed and optimized using gradient-based methods.

In this work, we present a thorough analysis of using the IB functional for training DNNs. Specifically, we show that in deterministic DNNs the IB functional leads to an ill-posed optimization problem by either being infinite for almost all parameter settings (Sec.~\ref{sec:infinity}) or by being a piecewise constant function of the parameters (Sec.~\ref{sec:pwconstant}). Moreover, we show in Sec.~\ref{sec:invariance} that the IB functional captures only a small subset of properties desirable for the representation $L$ when performing classification, and hence is not suitable as a cost function for training deterministic DNNs. We then show that the utility of the IB functional can partly be recovered by applying it for training stochastic DNNs or by including the decision rule (Sec.~\ref{subsec:IBsoldec} through Sec.~\ref{sec:solstochastic}). Furthermore, we argue in Sec.~\ref{sec:solreplace} that replacing the IB functional by a more well-behaved cost function inspired by~\eqref{eq:IB} may be an even better possibility. To utilize these considerations, we postpone discussing the related work until Sec.~\ref{sec:related}. We argue that the successes of~\cite{Kolchinsky_NLIB,Alemi_DVIB,Anonymous_ParametricIB, uncertaintyDVIB, vdb} must be attributed to and provide experimental evidence for the validity of these steps -- replacing the functional, making the DNN stochastic, including the decision rule -- and not on the fact that these works are based on the IB principle.

In our analysis, we make the uncommon assumption that the joint distribution between the features $X$ and the class $Y$ is known. This not only admits more rigorous statements, but makes them independent of the optimization heuristic used for training and corresponds to a best-case scenario for training. Nevertheless, we make regular comments on how our analysis changes in case only a finite dataset is available.

Finally, we wish to mention that while the focus of this work is on training DNNs, our results have strong implications on recent efforts in analyzing DNNs using the IB framework. Specifically, while statements about the trade-off between information-theoretic compression and preservation of class information may still be possible if the functional is finite, without applying the remedies from Sec.~\ref{sec:hotwo}, the IB framework does not admit making statements about the robustness, classification performance, or the representational simplicity of a given DNN.

\section{Setup and Preliminaries}\label{sec:notation}
We consider a feature-based classification problem. Suppose that the joint distribution between the $N$-dimensional (random) feature vector $X$ and the (random) class label $Y$ is given by $P_{X,Y}$. We denote realizations of random variables (RVs) with lower case letters, e.g., $X=x$ or $Y=y$. Unless otherwise specified, we assume that $X$ has an arbitrary distribution and that $Y$ has a discrete distribution on some finite set $\mathcal{Y}$ of class labels. 

Classification shall be performed by a feed-forward DNN. This is the same setup as the one discussed in \cite{Tishby_DLIB_ITW,Alemi_DVIB,Anonymous_IBTheory,Kolchinsky_NLIB,Tishby_BlackBox}. The DNN accepts the RV $X$ at the input and responds with the transformed RV $\lastlayer$, based on which the class label $Y$ can be estimated with a decision rule (see~\cite[Chap.~6.2]{deepbook} for a discussion on decision rules). Note that we use $\lastlayer$ to denote the output of the last layer of the DNN, e.g., it could be the output of the final softmax layer, rather than the estimate of $Y$ obtained after applying the decision rule. Hence $\lastlayer$ is a RV supported on (a subset of) $\reals^{|\lastlayer|}$ and not on the finite set $\mathcal{Y}$. If the DNN has $m$ hidden layers, then the column vector collecting all neuron outputs in the $i$-th hidden layer is the \emph{intermediate representation} denoted by $L_i$. We call $L_0\triangleq X$ the input and $L_{m+1}\triangleq \lastlayer$ the output of the DNN, while for $i\in\{1\dots m\}$, we call $L_i$ a latent representation. Abusing notation, $\left|L_i\right|$ denotes the number of neurons in the $i$-th layer, e.g., $|L_0|=N$. Whenever we talk about intermediate representations of which the layer number is immaterial we write $L$ instead of $L_i$.

If the DNN is deterministic, then $L_i$ and $L_{i+1}$ are related by a function $g_i{:}\ \mathbb{R}^{\left|L_i\right|}\to\mathbb{R}^{\left|L_{i+1}\right|}$ that maps the former to the latter and that depends on a set $\Theta_i$ of (weight and bias) parameters. E.g., if $\mathbf{W}_i$ is the matrix of weights between the $i$-th and $(i+1)$-th layer, $b_{i+1}$ the vector of biases for the $(i+1)$-th layer, and $\sigma{:}\ \mathbb{R}\to\mathbb{R}$ an activation function, then $\Theta_i=\{\mathbf{W}_{i},b_{i+1}\}$ and
\begin{equation}
 L_{i+1} = g_i(L_i) = \sigma\left(\mathbf{W}^T_{i}L_i +b_{i+1}  \right)
\end{equation}
where the activation function is applied coordinate-wise.
Whether the activation function is sigmoidal, ReLU, leaky ReLU, tanh, or softplus is immaterial for the results that follow, unless stated otherwise. We define $L_{i,j} \in \mathbb{R}$ to be the $j$-th component of the vector $L_i \in \mathbb{R}^{|L_i|}$ and $X_j = L_{0,j}$. We have $L_i=f_i(X)$, where $f_i \triangleq g_{i-1}\circ \cdots \circ g_1 \circ g_0$ shall be called the \emph{encoder for $L_i$}. Similarly, $\lastlayer=h_i(L_i)$, where $h_i\triangleq g_m\circ\cdots\circ g_i$ is called the \emph{decoder of $L_i$}. If the DNN is stochastic, then $g_i$ is a stochastic map parameterized by $\Theta_i$, and the encoder $f_i$ and decoder $h_i$ are obtained in the same way as for deterministic DNNs by appropriately concatenating the stochastic maps $\{g_i\}$.

We denote entropy, differential entropy, and mutual information by $\ent{\cdot}$, $\diffent{\cdot}$, and $\mutinf{\cdot;\cdot}$, respectively~\cite[Ch.~2~\&~9]{Cover_Information}. Specifically, if $X$ is continuous and $Y$ is discrete, then
\begin{equation}
 \mutinf{X;Y}=\ent{Y}-\ent{Y|X}=\diffent{X} - \diffent{X|Y}
\end{equation}
where all terms can be assumed to be finite. In contrast, $H(X)=\infty$ whenever $X$ is not discrete and $\diffent{Y}=-\infty$ whenever $Y$ is not continuous (see~\cite[Lemma~3.1~\&~p.~50]{Geiger_LossSystems}).

\section{Learning Representations for Classification}\label{sec:regularization}
The authors of~\cite{Tishby_DLIB_ITW} formulated  supervised deep learning as the goal of finding maximally compressed representations $\{L_i\}$ of the features $X$ that preserve as much information about the class variable $Y$ as possible. For classification tasks this goal is not sufficient. We will now present a list of properties of an intermediate representation $L$ desirable for the classification task. We do not accompany these properties with precise mathematical definitions -- this is out of scope of this paper and left for future work. Nevertheless, taken as guiding principles, these properties are sufficient to point out the shortcomings of the IB principle for training DNNs and to discuss ways to remedy them. For classification, the representation $L$ should
\begin{itemize}
 \item [P$1$] \textbf{inform about $Y$.} This means that the representation should contain as much information about the class variable $Y$ as was contained in the features $X$, i.e., $L$ should be a \emph{sufficient statistic} for $Y$.
 \item [P$2$] \textbf{be maximally compressed.} The representation $L$ should not tell more about $X$ than is necessary to correctly estimate $Y$, i.e., it should attain invariance, in some sense, to nuisance factors which are not relevant to the class label $Y$.  Compression, and consequently invariance, can, for example, be quantified statistically (e.g., $L$ is a \emph{minimal} sufficient statistic for $Y$) or geometrically (e.g., data points from different classes are mapped to different dense clusters in $\mathbb{R}^{|L|}$).
 \item [P$3$] \textbf{admit a simple decision function.} The successive intermediate representations should be such that the class $Y$ can be estimated from them using successively \enquote{simpler} functions. The term \enquote{simple} here has to be taken relative to the capabilities of the information sink or the system processing $L$. E.g., in DNNs, decisions are often made by searching for the output neuron with the maximum activation ($\arg\max$) or by binary quantization (for $|\mathcal{Y}|=2$) so the intermediate representation $\lastlayer$ should be such that these simple decision functions suffice to predict the class label from $\lastlayer \in \reals^{|\lastlayer|}$.
 \item [P$4$] \textbf{be robust.} This means that adding a small amount of noise to $X$ or transforming it with a well-behaved transform (e.g., affine transforms or small deformations) should not lead to big differences in the intermediate representation. E.g., the dense clusters in $\mathbb{R}^{|L|}$ corresponding to different classes should be far apart and the small deformations should not change the cluster in $L$ that a data point is mapped to.
\end{itemize}

Historically, the primary goals of training have been extracting information about $Y$ from the input $X$ such that this simplified extracted information can be effectively used by a simple decision mechanism to estimate $Y$ (P$1$ and P$3$). Traditionally, these goals have been achieved by using mean-squared error or cross-entropy as a cost function. Furthermore, robustness (P$4$) has been linked to improved generalization capabilities of learning algorithms \cite{Xurobust,novaksensitivity,zahavyrobust}; regularization measures such as dropout have been shown to instill robustness and improved generalization. Reference~\cite{Tishby_DLIB_ITW} has additionally introduced the idea of having maximally compressed intermediate representations (P$2$). The intuition behind this requirement is that this should avoid overfitting by making the network forget about the specific details of the individual examples and by making it invariant to nuisances not relevant for the classification task. 

In addition to achieving P$1$ through P$4$, one may wish that the DNN producing these intermediate representations is  architecturally economical. E.g., the DNN should consist of few hidden layers, of few neurons per layer, of few convolution filters or sparse weight matrices $\{\mathbf{W}_i\}$, or the inference process based on the DNN should be computationally economical. This goal becomes particularly important when deploying these DNNs on embedded/edge devices with limited computational resources and real-time processing constraints. While currently the network architectures leading to state-of-the-art performance in various classification tasks are highly over-parameterized, it has also been observed that a major portion of the network parameters can be pruned without significant deterioration in performance \cite{entropypruning,pruneconv,kairenprune} and it has been suggested that the over-parameterization of the network just provides ease of optimization during training \cite{lotteryticket}. Hence one may wish for obtaining certain desired characteristics in intermediate representations $L$ that either help in training architecturally/computationally economical DNNs to achieve state-of-the-art performance or that admit significant pruning after training without performance degradation.

Of course, these goals are not completely independent. For example, if a representation is robust and compressed, e.g., if the different regions in input domain $\reals^{N}$ corresponding to different classes are mapped to clusters dense and far apart in the intermediate representation domain $\reals^{|L|}$, then it may be easier to find a simple decision rule to estimate $Y$ from $L$. Such a representation $L$, however, may require an encoder $f$ with significant architectural/computational complexity.

Since goals P$1$-P$4$ are formulated as properties of the intermediate representation, achieving them can be accomplished by designing regularizers for $L$ based on, e.g., the joint distribution between $X$, $Y$, and $L$. Such regularization departs from classical regularization that depends on the parameters $\{\Theta_i\}$ of the DNN and relates closely to representation learning. Representation learning is an active field of research and various sets of desired properties for representations have previously been proposed. These are similar to our proposal but differ in subtle and key aspects. 

In \cite{bengiolearning}, Bengio et al.\ discussed desired characteristics of representations of the input in terms of invariant, disentangled and smoothly varying factors whereas our focus is on learning representations for a specific classification task. Nevertheless, P$1$-P$4$ have similarities to the properties discussed in \cite{bengiolearning}. For example, the hierarchical organization of explanatory factors discussed in \cite{bengiolearning} can lead to more abstract concepts at deeper layers. This subsequently may imply successively simpler decision functions required to estimate $Y$ from $L$ (P$3$). Similarly,~\cite{bengiolearning} discusses invariance and manifold learning mainly in the context of auto-encoders, focusing primarily on $X$. Our P$2$ goes one step further by including $Y$ in the picture, i.e., it aims to remove all information from $L$ that is not useful for determining $Y$. In a geometric understanding of compression this could mean to collapse the input manifolds corresponding to different class labels to, for example, separate dense clusters in $\reals^{|L|}$ (as observed in \cite[Fig.~2]{Alemi_DVIB} and~\cite[Fig.~2]{Kolchinsky_NLIB}).
Furthermore, in the context of representation learning, robustness is often related to denoising and contractive auto-encoders. However, our P$4$ aims to learn representations that are robust for the classification task, whereas for auto-encoders the aim is to learn robust representations to recover the input. 

The authors of~\cite{DisentangledRepresentations} focused on formulating a similar set of desired properties in terms of information-theoretic objectives. Their approach involved considering also network parameters as RVs, unlike \cite{Alemi_DVIB,Kolchinsky_NLIB,Tishby_DLIB_ITW} and our work where only $X$, $Y$, and latent representations (which are transformations of $X$) are RVs. Their definitions share similar intuitive meaning as ours; e.g., sufficiency is equivalent to P$1$, minimality and invariance follow the same spirit as P$2$, and invariance can also be partially linked to robustness (P$4$). However, as we discuss in Sec.~\ref{sec:IBfails} (at least for the case when only $X$ and $Y$ are RVs, but not the network parameters), defining P$2$-P$4$ in terms of information-theoretic quantities may not imply characteristics in DNNs that are desired for a classification task.

Both \cite{bengiolearning} and \cite{DisentangledRepresentations} have introduced an additional desired property of representations that they call disentanglement. In the context of classification, disentanglement is meant to complement invariance (P$2$ and P$4$ in our case). Invariance is achieved by keeping the robust features which are relevant to the classification task, whereas disentanglement requires making the extracted relevant features independent from one another (in the sense of total correlation \cite{DisentangledRepresentations} or some other metric). We have not included this property in our list for the following two reasons: First, disentangling features not necessarily improves classification or generalization performance. Second, features that are understandable for humans are not necessarily statistically independent (such as, e.g., size and weight of an object). We believe that more experiments are necessary to determine whether (and when) disentanglement, separated from other desirable properties, improves classification performance or human understandability of the internal representations. Thus, for now, disentanglement is not included in our list of desirable properties.

\section{Why and How IB Fails for Training Deterministic DNNs}\label{sec:IBfails}

In this section, we investigate the problem of learning an intermediate representation $L_i$ (which can also be $L_{m+1}=\lastlayer$) by a deterministic DNN with a given structure via minimizing the IB functional, i.e., we consider\footnote{Note that since the DNN is deterministic, we have $\mutinf{X;L_i}=\ent{L_i}$. The IB functional thus coincides with the ``deterministic'' IB functional proposed in~\cite{Strouse_DIB}.}
\begin{equation}\label{eq:IBOpt}
 \min_{\Theta_{0}, \dots, \Theta_{i-1}} \mutinf{X;L_i} - \beta\mutinf{Y;L_i}.
\end{equation}

The IB functional applied to DNNs therefore focuses on P$1$ and P$2$, defining them via the mutual information terms $\mutinf{Y;L}$ and $\mutinf{X;L}$, respectively. Such an approach has been proposed by~\cite{Tishby_DLIB_ITW,Tishby_BlackBox} and, subsequently, the IB framework has been suggested as a possible design principle for DNNs~\cite{Tishby_DLIB_ITW,Alemi_DVIB,Kolchinsky_NLIB, uncertaintyDVIB}. It was claimed that on this basis compressed, simple, and robust representations can be obtained (see~\cite[Fig.~2]{Alemi_DVIB} and~\cite[Fig.~2]{Kolchinsky_NLIB}).

Indeed, while the intermediate layers of a  DNN with good performance are characterized by a high $\mutinf{Y;L}$, they do not need to have small $\mutinf{X;L}$ (cf. \cite{iRevnet}), indicating that a small value of the IB functional is not necessary for good classification performance. Furthermore, since the IB framework was introduced to regularize intermediate representations rather than DNN parameters, a small value of $\mutinf{X;L}$ does not imply low architectural/computational complexity, as was empirically observed in \cite{Tishby_BlackBox} . Finally, small values of $\mutinf{X;L}$ do not relate causally to improved generalization performance, as it has been observed based on empirical evidence in~\cite{Anonymous_IBTheory}.

We show that applying the IB framework for training DNNs in this way suffers from two more major issues: The first issue is that, in many practically relevant cases, the IB functional is either equal to infinity or a piecewise constant function of the set of parameters $\{\Theta_i\}$. This either makes the optimization problem ill-posed or makes solving it difficult. We investigate these issues in Secs~\ref{sec:infinity} and~\ref{sec:pwconstant}. The second issue, which we investigate in Sec.~\ref{sec:invariance}, is connected to the invariance of mutual information under bijections and shows that focusing on goals P$1$ and P$2$ is not sufficient for a good classification system, at least when capturing P$1$ and P$2$ within the IB functional~\eqref{eq:IBOpt}. Specifically, we show that minimizing the IB functional~\eqref{eq:IBOpt} does not necessarily lead to classifiers that are robust (P$4$) or that allow using simple decision functions (P$3$).

\subsection{Continuous Features: The IB Functional is Infinite}\label{sec:infinity}

Solving~\eqref{eq:IBOpt} requires that the IB functional can be evaluated for a set of parameters $\{\Theta_i\}$. Since $Y$ is a discrete RV with finite support, the precision term $\mutinf{Y;L}\le\ent{Y}$ is finite and can be computed (at least in principle). Suppose now that the distribution of the features $X$ has an absolutely continuous component. Under this assumption, the following theorem shows that, for almost every non-trivial choice of $\{\Theta_i\}$, the IB functional is infinite and, hence, its optimization is ill-posed. The proof is deferred to Sec.~\ref{sec:proof}.

\begin{thm}\label{thm:infinity}
 Let $X=L_0$ be an $N$-dimensional RV, the distribution of which has an absolutely continuous component with a probability density function $f_X$ that is continuous on a compact set $\mathcal{X}$ in $\reals^N$. Consider a DNN as in the setup of Sec.~\ref{sec:notation}. Suppose that the activation function $\sigma$ is either bi-Lipschitz or continuously differentiable with strictly positive derivative. Then, for every $i=1,\dots,m$ and almost every choice of weight matrices $\mathbf{W}_{0},\dots,\mathbf{W}_{i}$, we have
\begin{equation}
 \mutinf{X;L_{i+1}}=\infty.
\end{equation}
\end{thm}

\tikzset{cross/.style={cross out, draw=black, minimum size=5*(#1-\pgflinewidth), inner sep=0pt, outer sep=0pt},cross/.default={1pt}}
\pgfplotsset{every tick label/.append style={font=\footnotesize}}
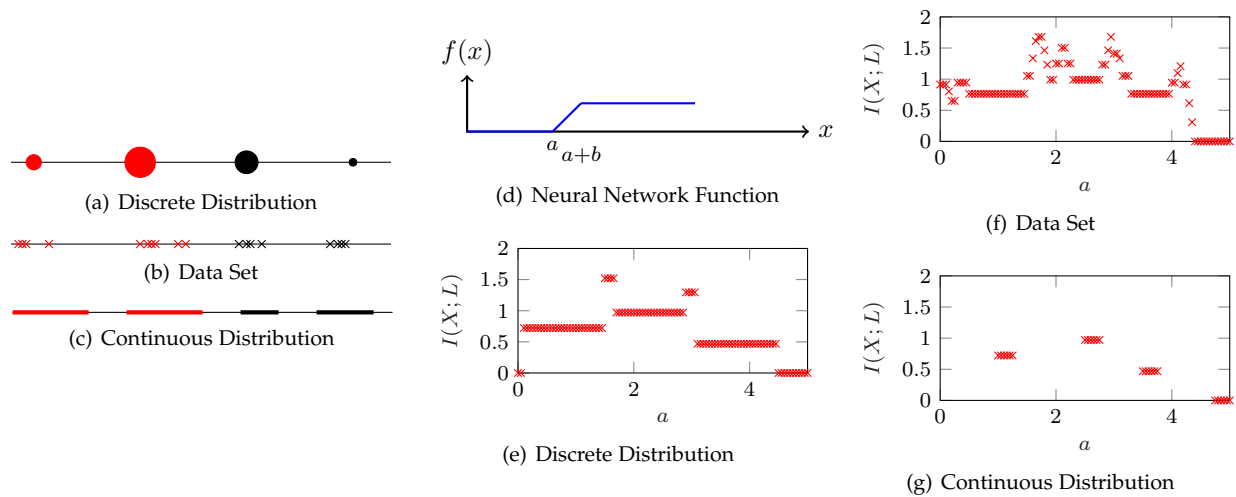
\begin{figure*}[t]
\centering 
	\begin{minipage}{0.3\textwidth}
		\subfigure[Discrete Distribution]{
		\begin{tikzpicture}
			\draw (0,0) -- (5,0);
			\filldraw[draw=red,fill=red] (0.3,0) circle (0.1);
			\filldraw[draw=red,fill=red] (1.7,0) circle (0.2);
			\filldraw[draw=black,fill=black] (3.1,0) circle (0.15);
			\filldraw[draw=black,fill=black] (4.5,0) circle (0.05);
		\end{tikzpicture}
		}\\
		\subfigure[Data Set]{
		\begin{tikzpicture}
			\draw (0,0) -- (5,0);
			\draw (0.1,0) node[cross,red] {};
			\draw (0.15,0) node[cross,red] {};
			\draw (0.2,0) node[cross,red] {};
			\draw (0.5,0) node[cross,red] {};
			\draw (1.7,0) node[cross,red] {};
			\draw (1.8,0) node[cross,red] {};
			\draw (1.85,0) node[cross,red] {};
			\draw (1.9,0) node[cross,red] {};
			\draw (2.2,0) node[cross,red] {};
			\draw (2.3,0) node[cross,red] {};
			\draw (3.0,0) node[cross,black] {};
			\draw (3.1,0) node[cross,black] {};
			\draw (3.15,0) node[cross,black] {};
			\draw (3.3,0) node[cross,black] {};
			\draw (4.2,0) node[cross,black] {};
			\draw (4.3,0) node[cross,black] {};
			\draw (4.35,0) node[cross,black] {};
			\draw (4.4,0) node[cross,black] {};
		\end{tikzpicture}
		}\\
		\subfigure[Continuous Distribution]{
		\begin{tikzpicture}
			\draw (0,0) -- (5,0);
			\draw[ultra thick,draw=red] (0,0) -- (1,0);
			\draw[ultra thick,draw=red] (1.5,0) -- (2.5,0);
			\draw[ultra thick,draw=black] (3,0) -- (3.5,0);
			\draw[ultra thick,draw=black] (4,0) -- (4.75,0);
		\end{tikzpicture}
		}\\
	\end{minipage}
	\begin{minipage}{0.3\textwidth}
	\subfigure[Neural Network Function]{
		\begin{tikzpicture}[scale=1.5]
			\draw [<->,thick] (0,0.5) node (yaxis) [above] {$f(x)$}
				|- (3,0) node (xaxis) [right] {$x$};
			\draw[ thick,draw=blue] (0,0) -- (0.75,0);
			\draw[ thick,draw=blue] (0.75,0) -- (1,0.25);
			\draw[ thick,draw=blue] (1,0.25) -- (2,0.25);
			\draw (0.75,0) node[below] {\footnotesize $a$};
			\draw (1,-0.05) node[below] {\footnotesize $a{+}b$};
		\end{tikzpicture}
		}
		\subfigure[Discrete Distribution]{
		\begin{tikzpicture}
		\begin{axis}[%
		width=1.5in,
		height=0.65in,
		scale only axis,
		xmin=0,
		xmax=5,
    ylabel near ticks,
    xlabel near ticks,
		xlabel style={font=\color{white!15!black}},
		xlabel={\footnotesize $a$},
		ymin=0,
		ymax=2,
		ylabel style={font=\color{white!15!black}},
		ylabel={\footnotesize $\mutinf{X;L}$},
		axis background/.style={fill=white}
		]
		\addplot [only marks, color=red, mark=x, forget plot,]
		table[row sep=crcr]{%
		0.000000 -0.000000\\
0.050000 -0.000000\\
0.100000 0.721928\\
0.150000 0.721928\\
0.200000 0.721928\\
0.250000 0.721928\\
0.300000 0.721928\\
0.350000 0.721928\\
0.400000 0.721928\\
0.450000 0.721928\\
0.500000 0.721928\\
0.550000 0.721928\\
0.600000 0.721928\\
0.650000 0.721928\\
0.700000 0.721928\\
0.750000 0.721928\\
0.800000 0.721928\\
0.850000 0.721928\\
0.900000 0.721928\\
0.950000 0.721928\\
1.000000 0.721928\\
1.050000 0.721928\\
1.100000 0.721928\\
1.150000 0.721928\\
1.200000 0.721928\\
1.250000 0.721928\\
1.300000 0.721928\\
1.350000 0.721928\\
1.400000 0.721928\\
1.450000 0.721928\\
1.500000 1.521928\\
1.550000 1.521928\\
1.600000 1.521928\\
1.650000 1.521928\\
1.700000 0.970951\\
1.750000 0.970951\\
1.800000 0.970951\\
1.850000 0.970951\\
1.900000 0.970951\\
1.950000 0.970951\\
2.000000 0.970951\\
2.050000 0.970951\\
2.100000 0.970951\\
2.150000 0.970951\\
2.200000 0.970951\\
2.250000 0.970951\\
2.300000 0.970951\\
2.350000 0.970951\\
2.400000 0.970951\\
2.450000 0.970951\\
2.500000 0.970951\\
2.550000 0.970951\\
2.600000 0.970951\\
2.650000 0.970951\\
2.700000 0.970951\\
2.750000 0.970951\\
2.800000 0.970951\\
2.850000 0.970951\\
2.900000 1.295462\\
2.950000 1.295462\\
3.000000 1.295462\\
3.050000 1.295462\\
3.100000 0.468996\\
3.150000 0.468996\\
3.200000 0.468996\\
3.250000 0.468996\\
3.300000 0.468996\\
3.350000 0.468996\\
3.400000 0.468996\\
3.450000 0.468996\\
3.500000 0.468996\\
3.550000 0.468996\\
3.600000 0.468996\\
3.650000 0.468996\\
3.700000 0.468996\\
3.750000 0.468996\\
3.800000 0.468996\\
3.850000 0.468996\\
3.900000 0.468996\\
3.950000 0.468996\\
4.000000 0.468996\\
4.050000 0.468996\\
4.100000 0.468996\\
4.150000 0.468996\\
4.200000 0.468996\\
4.250000 0.468996\\
4.300000 0.468996\\
4.350000 0.468996\\
4.400000 0.468996\\
4.450000 0.468996\\
4.500000 -0.000000\\
4.550000 -0.000000\\
4.600000 -0.000000\\
4.650000 -0.000000\\
4.700000 -0.000000\\
4.750000 -0.000000\\
4.800000 -0.000000\\
4.850000 -0.000000\\
4.900000 -0.000000\\
4.950000 -0.000000\\
5.000000 -0.000000\\
		};
		\end{axis}
		\end{tikzpicture}%
		}
	\end{minipage}
	\begin{minipage}{0.3\textwidth}
		\subfigure[Data Set]{
		\begin{tikzpicture}
		\begin{axis}[%
		width=1.5in,
		height=0.65in,
		scale only axis,
		xmin=0,
		xmax=5,
    ylabel near ticks,
    xlabel near ticks,
		xlabel style={font=\color{white!15!black}},
		xlabel={\footnotesize $a$},
		ymin=0,
		ymax=2,
		ylabel style={font=\color{white!15!black}},
		ylabel={\footnotesize $\mutinf{X;L}$},
		axis background/.style={fill=white}
		]
		\addplot [only marks, color=red, mark=x, forget plot,]
		table[row sep=crcr]{%
		0.000000 0.914183\\
0.050000 0.914183\\
0.100000 0.914183\\
0.150000 0.803072\\
0.200000 0.650022\\
0.250000 0.650022\\
0.300000 0.944489\\
0.350000 0.944489\\
0.400000 0.944489\\
0.450000 0.944489\\
0.500000 0.764205\\
0.550000 0.764205\\
0.600000 0.764205\\
0.650000 0.764205\\
0.700000 0.764205\\
0.750000 0.764205\\
0.800000 0.764205\\
0.850000 0.764205\\
0.900000 0.764205\\
0.950000 0.764205\\
1.000000 0.764205\\
1.050000 0.764205\\
1.100000 0.764205\\
1.150000 0.764205\\
1.200000 0.764205\\
1.250000 0.764205\\
1.300000 0.764205\\
1.350000 0.764205\\
1.400000 0.764205\\
1.450000 0.764205\\
1.500000 1.052941\\
1.550000 1.052941\\
1.600000 1.335506\\
1.650000 1.611383\\
1.700000 1.679429\\
1.750000 1.679429\\
1.800000 1.462755\\
1.850000 1.232660\\
1.900000 0.991076\\
1.950000 0.991076\\
2.000000 1.251629\\
2.050000 1.251629\\
2.100000 1.503258\\
2.150000 1.503258\\
2.200000 1.251629\\
2.250000 1.251629\\
2.300000 0.991076\\
2.350000 0.991076\\
2.400000 0.991076\\
2.450000 0.991076\\
2.500000 0.991076\\
2.550000 0.991076\\
2.600000 0.991076\\
2.650000 0.991076\\
2.700000 0.991076\\
2.750000 0.991076\\
2.800000 1.232660\\
2.850000 1.232660\\
2.900000 1.462755\\
2.950000 1.679429\\
3.000000 1.410848\\
3.050000 1.410848\\
3.100000 1.335506\\
3.150000 1.052941\\
3.200000 1.052941\\
3.250000 1.052941\\
3.300000 0.764205\\
3.350000 0.764205\\
3.400000 0.764205\\
3.450000 0.764205\\
3.500000 0.764205\\
3.550000 0.764205\\
3.600000 0.764205\\
3.650000 0.764205\\
3.700000 0.764205\\
3.750000 0.764205\\
3.800000 0.764205\\
3.850000 0.764205\\
3.900000 0.764205\\
3.950000 0.764205\\
4.000000 0.944489\\
4.050000 0.944489\\
4.100000 1.097538\\
4.150000 1.208649\\
4.200000 0.914183\\
4.250000 0.914183\\
4.300000 0.614369\\
4.350000 0.309543\\
4.400000 -0.000000\\
4.450000 -0.000000\\
4.500000 -0.000000\\
4.550000 -0.000000\\
4.600000 -0.000000\\
4.650000 -0.000000\\
4.700000 -0.000000\\
4.750000 -0.000000\\
4.800000 -0.000000\\
4.850000 -0.000000\\
4.900000 -0.000000\\
4.950000 -0.000000\\
5.000000 -0.000000\\
		};
		\end{axis}
		\end{tikzpicture}%
		}
\subfigure[Continuous Distribution]{
		\begin{tikzpicture}
		\begin{axis}[%
		width=1.5in,
		height=0.65in,
		scale only axis,
		xmin=0,
		xmax=5,
    ylabel near ticks,
    xlabel near ticks,
		xlabel style={font=\color{white!15!black}},
		xlabel={\footnotesize $a$},
		ymin=0,
		ymax=2,
		ylabel style={font=\color{white!15!black}},
		ylabel={\footnotesize $\mutinf{X;L}$},
		axis background/.style={fill=white}
		]
		\addplot [only marks, color=red, mark=x, forget plot,]
		table[row sep=crcr]{%
1.000000 0.721928\\
1.050000 0.721928\\
1.100000 0.721928\\
1.150000 0.721928\\
1.200000 0.721928\\
1.250000 0.721928\\
2.500000 0.970951\\
2.550000 0.970951\\
2.600000 0.970951\\
2.650000 0.970951\\
2.700000 0.970951\\
2.750000 0.970951\\
3.500000 0.468996\\
3.550000 0.468996\\
3.600000 0.468996\\
3.650000 0.468996\\
3.700000 0.468996\\
3.750000 0.468996\\
4.750000 -0.000000\\
4.800000 -0.000000\\
4.850000 -0.000000\\
4.900000 -0.000000\\
4.950000 -0.000000\\
5.000000 -0.000000\\
		};
		\end{axis}
		\end{tikzpicture}%
		}
	\end{minipage}
    \caption{(a)-(c): The line segment depicts the set $[0,1]$, from which the feature RV $X$ takes its values. Red (black) color indicates feature values corresponding to class $Y=0$ ($Y=1$). (a): The one-dimensional feature variable has a discrete distribution with mass points as indicated by the circles. The size of the circles is proportional to the probability mass. (b): Training based on a data set $\mathcal{D}$. Crosses indicate data points. (c): The one-dimensional feature variable has a continuous distribution with support indicated by the thick lines, the probability masses on each interval are identical to the probability masses of the points in (a). (d): The function $f$ implemented by a DNN with a one hidden layer with two neurons, ReLU activation functions, and a single output neuron. The parameters leading to this function are  $\Theta_0=\{[1; 1],[-a; -a-b]\}$ and $\Theta_1=\{[1,-1],0\}$. (e)-(g) show the mutual information $\mutinf{X;L}$ as a function of the parameter $a$, for $b=0.25$, evaluated on a grid of $a$ ranging from 0 to 5 in steps of 0.05. It can be seen that the mutual information is piecewise constant. The missing values in (g) indicate that the mutual information is infinite at the respective positions.}
    \label{fig:piecewise} 
\end{figure*}

In~\cite[Appendix~C]{Anonymous_IBTheory} it has been observed that the  mutual information between the continuously distributed input $X$ and an intermediate representation $L$ becomes infinite if $L$ has a continuous distribution. This assumption is  often not satisfied: For example, the output of a ReLU activation function is, in general, the mixture of a continuous and a discrete distribution. Also, if the number of neurons $|L_i|$ of some layer exceeds the number of neurons of any preceding layer or the dimension of the input $X$, then $L_i$ cannot have a continuous distribution on $\reals^{|L_i|}$ if the activation functions satisfy the conditions of Theorem~\ref{thm:infinity}. Therefore, our Theorem~\ref{thm:infinity} is more general than~\cite[Appendix~C]{Anonymous_IBTheory} in the sense that continuity of the distribution $L$ is not required.

Theorem~\ref{thm:infinity} shows that the IB functional leads to an ill-posed optimization problem for, e.g., sigmoidal and tanh activation functions (which are continuously differentiable with strictly positive derivative) as well as for leaky ReLU activation functions (which are bi-Lipschitz). The situation is different for ReLU or step activation functions. For these activation functions, the intermediate representations $L$ may have purely discrete distributions, from which follows that the IB functional is finite (at least for a non-vanishing set of parameters). As we discuss in Sec.~\ref{sec:pwconstant}, in such cases other issues dominate, such as the IB functional being piecewise constant.

Note that the issue discussed in this section is not that the IB functional is difficult to compute, as was implied in \cite{Alemi_DVIB,Kolchinsky_NLIB}. Indeed, Theorem~\ref{thm:infinity} provides us with the correct value of the IB functional, i.e., infinity, for almost every choice of weight matrices. At the same time, Theorem~\ref{thm:infinity} shows that in such a scenario it is ill-advised to \emph{estimate} mutual information from a data sample, as such estimators are only valid if the true mutual information determined by the assumed underlying distribution is finite. Indeed, the estimate $\estmutinf{X;L}$ reveals more about the estimator and the dataset than it does about the true mutual information, as the latter is always infinite by Theorem~\ref{thm:infinity}; see also the discussion in~\cite[Sec.~2~\&~Appendix~C]{Anonymous_IBTheory}

\subsection{Discrete Features or Learning from Data: The IB Functional is Piecewise Constant}\label{sec:pwconstant}

We next assume that the features have a discrete distribution, i.e., $X$ can assume only a finite number of different points in $\dom{X}\subset\reals^N$. For example, one may assume that $X$ is a RV over black-and-white images with $N$ pixels, in which case the distribution of $X$ is supported on $\dom{X}=\{0,1\}^N$. In such a case, the entropy of $X$ is finite and, thus, so is the entropy of every intermediate representation $L$. More precisely, since the DNN is deterministic, the distribution of $L$ is discrete as well, from which follows that $\mutinf{X;L}=\ent{L}$ can assume only finitely many values. Similarly, since both $L$ and $Y$ are discrete, also $\mutinf{Y;L}$ can assume only finitely many different values. Indeed, $\mutinf{X;L}$ and $\mutinf{Y;L}$ may change only when two different $x \in \dom{X}$ that were previously mapped to different values of intermediate representation now get mapped to the same value or vice-versa. As a consequence, the IB functional is a piecewise constant function of the parameters $\{\Theta_i\}$ and, as such, difficult to optimize. Specifically, the gradient of the IB functional w.r.t.\ the parameter values is zero almost everywhere, and one has to resort to other optimization heuristics that are not gradient-based.

The problem of piecewise continuity persists if the empirical joint distribution of $X$ and $Y$ based on a dataset $\mathcal{D}$ with finitely many data points is used to optimize the IB objective. The entropy of $X$ equals $\log|\mathcal{D}|$, and the IB functional remains piecewise constant. Indeed, $\mutinf{X;L}$ and $\mutinf{Y;L}$ may change only when two different data points which were previously mapped to different values of intermediate representation now get mapped to the same value or vice-versa. It was shown empirically in~\cite[Fig.~15]{Anonymous_IBTheory} that $\mutinf{X;L}=\log|\mathcal{D}|$ throughout training, i.e., for a large selection of weight matrices.

Finally, the IB functional can be piecewise constant also for a continuously distributed feature RV $X$ if step or ReLU activation functions are used. This can happen, for example, if the distribution of $X$ is supported on a disconnected set $\dom{X}\subset\reals^N$. Such a situation is depicted in Fig.~\ref{fig:piecewise} together with the scenarios of a discretely distributed feature RV $X$ and a dataset $\mathcal{D}$.

\subsection{Invariance under Bijections: The IB Functional is Insufficient}\label{sec:invariance}
Leaving aside the fundamental problems discussed in Sec.~\ref{sec:infinity} and Sec.~\ref{sec:pwconstant}, we now show that the IB functional is insufficient to fully characterize classification problems using DNNs. Specifically, we show that training a DNN by minimizing~\eqref{eq:IBOpt} does not lead to representations that admit simple decision functions (P$3$) or are robust to noise, well behaved transformations or small distortions (P$4$). To this end, we give several examples comparing two DNNs whose intermediate representations are equivalent in terms of the IB functional, but where one of them is clearly a more desirable solution. Since the IB functional does not give preference to any of the two solutions, we conclude that it is insufficient to achieve intermediate representations satisfying the requirements stated in Sec.~\ref{sec:regularization}. For the sake of argument, we present simple, synthetic examples instead of empirical evidence on real-world datasets to illustrate these shortcomings. On the one hand, the examples have all the essential aspects associated with training a DNN for a practical classification task. On the other hand, because of the simplicity of the examples, they lend themselves to clearly highlighting and explaining different shortcomings in isolation. One can then easily extrapolate how one may encounter these issues in practical scenarios.

\begin{figure*}[p!]
	\centering 
	\begin{minipage}{0.6\textwidth}
		 \begin{tikzpicture}
		\begin{scope} 
		\begin{axis}[
		ylabel near ticks,
		xlabel near ticks,
		ylabel={$X_2$},
		xlabel={$X_1$},
		xmin=0, xmax=8.5,
		ymin=0, ymax=2,
		width=0.8\textwidth,height=0.16\textheight
		]
		\node at (172,110) {\tikz[cm={-1,0,0,0.5,(0,0)}]\asymcloud[.044]{fill=red!20};}; 
		\node at (175,140) {$\mathcal{S}_1$};
		\node at (372,25) {\tikz[cm={-1,0,0,0.5,(0,0)}]\asymcloudd[.057]{fill=black!20};}; 
		\node at (372,50) {$\mathcal{S}_2$};
		\node at (540,35) {\tikz[cm={-1,0,0,0.5,(0,0)}]\asymclouddd[.03]{fill=red!20};}; 
		\node at (540,55) {$\mathcal{S}_3$};
		\node at (786,120) {\tikz[cm={-1,0,0,0.5,(0,0)}]\asymcloudddd[.0685]{fill=black!20};}; 
		\node at (785,150) {$\mathcal{S}_4$};
		\end{axis}
		\end{scope}
		
		\begin{scope}[yshift = -3.5cm] 
		\begin{axis}[
		ylabel near ticks,
		xlabel near ticks,
		ylabel={$f'_{disc}$},
		xlabel={$X_1$},
		xmin=0, xmax=8.5,
		ymin=0, ymax=1,
		width=0.8\textwidth,height=0.16\textheight
		]
		\addplot[blue, mark=+,thick] file {figures/examples/Y1disc.txt};
		\end{axis}
		\end{scope}
		
		\begin{scope}[yshift = -7cm] 
		\begin{axis}[
		ylabel near ticks,
		xlabel near ticks,
		ylabel={$f''_{disc}$},
		xlabel={$X_1$},
		xmin=0, xmax=8.5,
		ymin=0, ymax=1,
		width=0.8\textwidth,height=0.16\textheight
		]
		\addplot[blue, mark=+,thick] file {figures/examples/Y2disc.txt};
		\end{axis}
		\end{scope}
		
		\begin{scope}[yshift = -10.5cm] 
		\begin{axis}[
		ylabel near ticks,
		xlabel near ticks,
		ylabel={$f'_{cont}$},
		xlabel={$X_1$},
		xmin=0, xmax=8.5,
		ymin=0, ymax=1,
		ytick={0.25,0.5,0.75,1},
		width=0.8\textwidth,height=0.16\textheight
		]
		\addplot[blue, mark=+,thick] file {figures/examples/Y1cont.txt};
		\end{axis}
		\end{scope}
		
		\begin{scope}[yshift = -14cm] 
		\begin{axis}[
		ylabel near ticks,
		xlabel near ticks,
		ylabel={$f''_{cont}$},
		xlabel={$X_1$},
		xmin=0, xmax=8.5,
		ymin=0, ymax=1,
			ytick={0.25,0.5,0.75,1},
		width=0.8\textwidth,height=0.16\textheight
		]
		\addplot[blue, mark=+,thick] file {figures/examples/Y2cont.txt};
		\end{axis}
		\end{scope}
		
		\begin{scope}[yshift = -17.5cm] 
		\begin{axis}[
		ylabel near ticks,
		xlabel near ticks,
		ylabel={$f^{'''}_{disc}$},
		xlabel={$X_1$},
		xmin=0, xmax=8.5,
		ymin=0, ymax=1,
		width=0.8\textwidth,height=0.16\textheight
		]
		\addplot[blue, mark=+,thick] file {figures/examples/Y3disc.txt};
		\end{axis}
		\end{scope}
		
		\begin{scope}[yshift = -21cm] 
		\begin{axis}[
		ylabel near ticks,
		xlabel near ticks,
		ylabel={$f^{'''}_{cont}$},
		xlabel={$X_1$},
		xmin=0, xmax=8.5,
		ymin=0, ymax=1,
			ytick={0.25,0.5,0.75,1},
		width=0.8\textwidth,height=0.16\textheight
		]
		\addplot[blue, mark=+,thick] file {figures/examples/Y3cont.txt};
		\end{axis}
		\end{scope}
   
		\draw[dashed](1.01,-21) -- (1.01,2.4);
		\draw[dashed](1.937,-21) -- (1.937,2.4);
		\draw[dashed](2.525,-21) -- (2.525,2.4);
		\draw[dashed](3.62,-21) -- (3.62,2.4);
		\draw[dashed](4.21,-21) -- (4.21,2.4);
		\draw[dashed](4.8,-21) -- (4.8,2.4);
		\draw[dashed](5.895,-21) -- (5.895,2.4);
		\draw[dashed](7.07,-21) -- (7.07,2.4);
	 \end{tikzpicture}	
	\end{minipage}
  \begin{minipage}{0.3\textwidth}
  	\begin{tikzpicture}[yscale=0.5]
 		\node at (4,0) {};
   \begin{scope}[yshift=-12cm] 
   \draw[] (4,3) -- node[below] {$\lastlayer'_{disc}$} (9,3);
   \filldraw[draw=red,fill=red] (5,3) circle (0.075);
   \filldraw[draw=red,fill=red] (6,3) circle (0.075);
   \filldraw[draw=black,fill=black] (7,3) circle (0.075);
   \filldraw[draw=black,fill=black] (8,3) circle (0.075);
   \end{scope}
   
   \begin{scope}[yshift=-19cm] 
   \draw[] (4,3) -- node[below] {$\lastlayer''_{disc}$} (9,3);
   \filldraw[draw=red,fill=red] (5,3) circle (0.075);
   \filldraw[draw=black,fill=black] (6,3) circle (0.075);
   \filldraw[draw=red,fill=red] (7,3) circle (0.075);
   \filldraw[draw=black,fill=black] (8,3) circle (0.075);
   \end{scope}
   
   \begin{scope} [yshift=-26cm] 
   \draw[] (4,3) -- node[below] {$\lastlayer'_{cont}$} (9,3);
   \draw[ultra thick,draw=red] (4,3) -- (6.5,3);
   \draw[ultra thick,draw=black] (6.5,3) -- (9,3);
   \end{scope}
   
   \begin{scope} [yshift=-33cm] 
   \draw[] (4,3) -- node[below] {$\lastlayer^{''}_{cont}$} (9,3);
   \draw[ultra thick,draw=red] (4,3) -- (5.25,3);
   \draw[ultra thick,draw=black] (5.25,3) -- (6.5,3);
   \draw[ultra thick,draw=red] (6.5,3) -- (7.75,3);
   \draw[ultra thick,draw=black] (7.75,3) -- (9,3);
   \end{scope}
   
   \begin{scope}[yshift=-40cm] 
   \draw[] (4,3) -- node[below] {$\lastlayer^{'''}_{disc}$} (9,3);
   \filldraw[draw=red,fill=red] (5,3) circle (0.075);
   \filldraw[draw=red,fill=red] (6,3) circle (0.075);
   \filldraw[draw=black,fill=black] (7,3) circle (0.075);
   \filldraw[draw=black,fill=black] (8,3) circle (0.075);
   \end{scope}
   
   \begin{scope} [yshift=-47cm] 
   \draw[] (4,3) -- node[below] {$\lastlayer^{'''}_{cont}$} (9,3);
   \draw[ultra thick,draw=red] (4,3) -- (5.25,3);
   \draw[ultra thick,draw=black] (7.75,3) -- (9,3);
   \end{scope}
   \end{tikzpicture}
\end{minipage}
	\caption{Representational simplicity and robustness in binary classification: The top figure on the L.H.S. illustrates the two-dimensional input space and the support of the input $X$ in $\mathbb{R}^2$. The rest of the figures on L.H.S. show various functions of $X_1$ (since $Y$ only depends on $X_1$) implementable using a ReLU-based DNN. The figures on R.H.S. show the output RVs $\lastlayer$ when $X$ is transformed via the corresponding functions on the L.H.S. Red (black) color indicates feature values corresponding to class $Y=0$ ($Y=1$).}
	\label{fig:problem} 
\end{figure*}
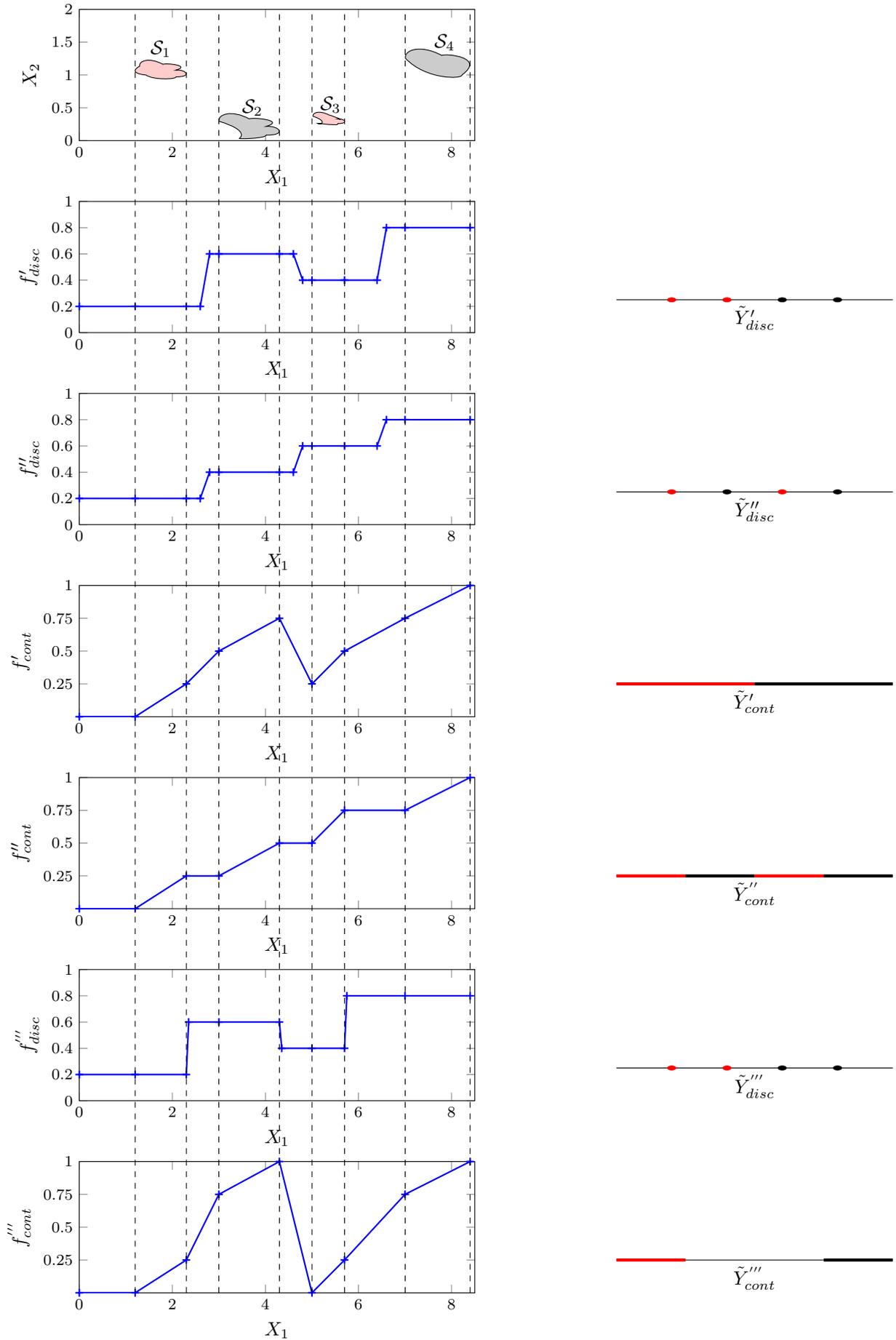

We consider a binary classification problem (i.e., $Y\in\{0,1\}$) based on two-dimensional input $X$ shown in Fig~\ref{fig:problem}. The input RV/samples take values in the four disjoint compact sets $\{\dom{S}_1, \dom{S}_2,\dom{S}_3,\dom{S}_4\}$, $\dom{S}_1$ and $\dom{S}_3$ (marked red) corresponding to $Y=0$ and $\dom{S}_2$ and $\dom{S}_4$ (marked black) corresponding to $Y=1$. We also define $\dom{T} = \dom{S}_1\bigcup \dom{S}_2\bigcup \dom{S}_3\bigcup \dom{S}_4$. Perfect classification is possible in principle, i.e., the distributions of $X$ given different classes have disjoint support\footnote{We refer the reader to~\cite{Kolchinsky_Pathologies} regarding additional issues under which the IB framework suffers in this scenario.} and $\mutinf{X;Y}=\ent{Y}$. The DNNs required to obtain the intermediate representations discussed in the examples can be easily implemented using ReLU activation functions (the examples can be modified to work with other activation functions). For the sake of simplicity, in this example the class label depends only on one dimension of the input.
Moreover, in the examples we evaluate the IB functional for the output $\lastlayer$. The considerations are equally valid if the presented functions are encoders for a latent representation instead of $\lastlayer$. Finally, one can extend the examples where the intermediate representation has more than one dimension.

First, consider the two functions $f'_{disc}(\cdot)$ and $f''_{disc}(\cdot)$ on the left-hand side (L.H.S.) in Fig.~\ref{fig:problem}, implemented by two DNNs. The corresponding figures on the right-hand side (R.H.S.) show the support of the distribution of $\lastlayer'_{disc}$ and $\lastlayer''_{disc}$, i.e., to which $X$ is mapped by $f'_{disc}$ and $f''_{disc}$, respectively.  It is easy to see that the IB functional evaluates to the same value for both DNNs. Indeed, both DNNs have identical compression terms, i.e., $\mutinf{X;\lastlayer'_{disc}}=\mutinf{X;\lastlayer''_{disc}}$ and perfect precision, i.e., $\mutinf{Y;\lastlayer'_{disc}}=\mutinf{Y;\lastlayer''_{disc}}=\ent{Y}$. However, while $f'_{disc}$ admits a simple decision by thresholding $\lastlayer'_{disc}$ at $1/2$, the representation $\lastlayer''_{disc}$ requires a more elaborate decision rule. This holds true regardless if the input $X$ has a continuous or discrete distribution supported on a subset of $\dom{T}$. It also holds if the computations are done based on a dataset with input samples lying in $\dom{T}$.

The same phenomenon can be observed when we compare the two functions $\lastlayer'_{cont}=f'_{cont}(X_1)$ and $\lastlayer''_{cont}=f''_{cont}(X_1)$ in Fig.~\ref{fig:problem}, implemented by two DNNs. If the input has a continuous distribution supported on $\dom{T}$, this leads to the continuous output RVs $\lastlayer'_{cont}$ and $\lastlayer''_{cont}$ shown on the R.H.S. Again both DNNs have perfect precision and identical compression terms, where $\mutinf{X;\lastlayer'_{cont}}$ and $\mutinf{X;\lastlayer''_{cont}}$ are both infinite in this case due to a continuously distributed $\lastlayer'_{cont}$ and $\lastlayer''_{cont}$. However,  $f'_{cont}$ admits a simple decision by thresholding $\lastlayer'_{cont}$ at $1/2$, whereas $\lastlayer''_{cont}$ requires a more elaborate decision rule. 

\begin{figure}[t]
	\begin{center}
		\begin{tikzpicture}
		
		\draw[] (0,0) -- (0,3);
		\draw[] (0,0) -- (3,0);
		
		\draw[ultra thick,draw=black] (0,1.5) -- (0,3);
		\draw[ultra thick,draw=red] (0,0) -- (1.5,0);
		
		\draw[dashed] (-0.25,1.75)  --  (1.75,-.25) node[right] {$\Theta_0^I$};
		\draw[dotted] (-0.25,0.75) -- (2.25,0.75) node[right] {$\Theta_0^{II}$};
		
		\filldraw[draw=blue,fill=blue] (0.1,1.2) circle (0.075);
		
		\draw[] (4,2) -- node[below] {$\lastlayer^I=f^I(X)$} (7,2);
		\draw[ultra thick,draw=red] (4,2) -- (5.5,2);
		\draw[ultra thick,draw=black] (5.5,2) -- (7,2);
		\filldraw[draw=blue,fill=blue] (5.3,2) circle (0.075);
		
		\draw[] (4,0.5) -- node[below] {$\lastlayer^{II}=f^{II}(X)$} (7,0.5);
		\filldraw[draw=red,fill=red] (4,0.5) circle (0.075);
		\draw[ultra thick,draw=black] (5.5,0.5) -- (7,0.5);
		\filldraw[draw=blue,fill=blue] (5.2,0.5) circle (0.075);
		
		\end{tikzpicture}
		\caption{Robustness in binary classification. The L.H.S.\ shows the feature space, with $X_1$ on the horizontal and $X_2$ on the vertical axis. One can see that $X$ is distributed on $[0,1/2]\times[0,\varepsilon]$ if $Y=0$ and on $[0,\varepsilon]\times[1/2,1]$ if $Y=1$, for $0<\varepsilon\ll1$. The R.H.S.\ shows the supports of the distributions $P_{\lastlayer|Y=0}$ and $P_{\lastlayer|Y=1}$, obtained by two different DNNs with identical IB functionals. The blue dot represents a noisy feature or a data point not in the training set. See text for details.}
		\label{fig:robustness}
	\end{center}
	\vskip -0.2in
\end{figure}
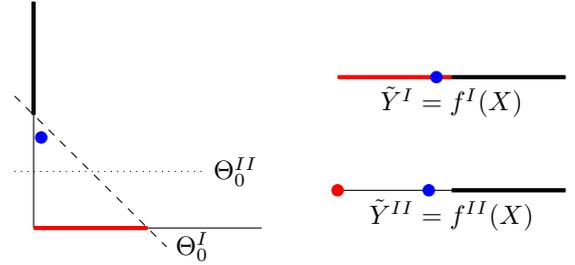

We finally turn to the question of robustness against noisy inputs. This, in general, cannot be answered by looking at the intermediate representations alone as we show in the following two examples. To this end, first consider the situation depicted in Fig.~\ref{fig:robustness}. As it can be seen, perfect classification is possible with a single neuron with a ReLU activation function. We consider two different DNNs with no hidden layers and single output neuron with parameterizations $\Theta_0^I=\{[1,1],0\}$ and $\Theta_0^{II}=\{[0,1],0\}$. Both parameterizations are equivalent in terms of the IB functional, leading to identical precision and compression terms. Note, however, that the DNN $f^{II}$ parameterized by $\Theta_0^{II}$ is more robust to small amount of noise or distortion than $f^I$. This can be seen by the blue dot in Fig.~\ref{fig:robustness} indicating a noisy input generated (with high probability) by class label $Y=1$. While $f^I$ does not admit distinguishing this point from features generated by class label $Y=0$, $f^{II}$ does (see R.H.S.\ of Fig.~\ref{fig:robustness}). Indeed, thresholding $\lastlayer^I$ at $0.5$ and $\lastlayer^{II}$ at $0.25$ yields the decision regions indicated by dashed and dotted lines on the left of Fig.~\ref{fig:robustness}.

As second example, consider the two DNNs implementing the functions $f'_{disc}$ and $f'''_{disc}$ in Fig.~\ref{fig:problem}. The corresponding RVs $\lastlayer'_{disc}$ and $\lastlayer'''_{disc}$ for the given $X$ are shown on the R.H.S.\ respectively. For the given distribution of $X$, we notice that $\lastlayer'_{disc} = \lastlayer'''_{disc}$, i.e., the two DNNs implement the same function over the support of $X$. Adding noise to $X$ or distorting $X$ has the potential to enlarge the support of its distribution. In this case, $f'_{disc}$ will be more robust to such noise and distortions when compared to $f'''_{disc}$, due to the sharp transitions of $f'''_{disc}$ outside the current support of the distribution of $X$.  This holds true whether the input $X$ is continuous or discrete with support over a subset of $\dom{T}$. It also holds if the computations are done based on a dataset with input samples lying in $\dom{T}$. 

In conclusion, the IB framework may serve to train a DNN whose output is maximally compressed (in an information-theoretic sense) and maximally informative about the class. However, one cannot expect that the DNN output admits taking a decision with a simple function or that the DNN is robust against noisy or distorted features. This holds true also for DNNs with discrete-valued features $X$, such as those discussed  in \cite{Tishby_DLIB_ITW}, for which the issue of Sec.~\ref{sec:infinity} does not appear. Based on the link between robustness and generalization, this suggests that the IB functional also cannot be used to quantify generalization of a DNN during training. 

\section{How to Use IB-Like Cost Functions for Training DNNs}\label{sec:hotwo}
The issues we discussed in Sec.~\ref{sec:IBfails} apply to training deterministic DNNs. In this section we discuss possible remedies for these problems, such as forcing the intermediate representation to be discrete, training stochastic DNNs, and replacing the IB functional by a more well-behaved cost function inspired by the IB framework. With these approaches one can guarantee that the IB functional is finite and that specific pairs of intermediate representations, related by invertible transforms, are not equivalent anymore. However, some of the proposed approaches lead to the IB functional being piecewise constant, similar to the scenario in Sec.~\ref{sec:pwconstant}.

\subsection{Including the Decision Rule}\label{subsec:IBsoldec}
One approach to successfully apply the IB framework for DNN training is to include a simple decision rule. Specifically when applied to the output representation $\lastlayer$, for a fixed function $\delta{:}\ \mathbb{R}^{|\lastlayer|}\to\mathcal{Y}$ and $\outputguess=\delta(\lastlayer)$, the goal shall be to solve
\begin{equation}\label{eq:IBOptRule}
 \min_{\{\Theta_i\}} \mutinf{X;\outputguess} - \beta\mutinf{Y;\outputguess}.
\end{equation}
For example, for binary $Y$ and a single output neuron, one could set $\outputguess=\mathbb{I}[\lastlayer>0.5]$, where $\mathbb{I}[\cdot]$ is the indicator function; for $|\mathcal{Y}| >2$ output neurons, $\outputguess$ could be the index of the output neuron with the maximum value. 

Since $\outputguess$ has a discrete distribution with as many mass points as the class variable, the compression term $\mutinf{X;\outputguess}$ appears useless: compression is enforced by including the decision rule $\delta$. Similarly, the simplicity of the representation is automatically enforced by the simplicity of the decision function $\delta$ when solving~\eqref{eq:IBOptRule}. Moreover, the IB functional becomes computable for the output layer because we have $\mutinf{X;\outputguess} \le \ent{\outputguess}\le \log|\mathcal{Y}|$. Finally, for the example depicted in Fig.~\ref{fig:problem}, assuming that $\Prob{X\in\dom{S}_i} = \frac{1}{4}$ for $i  \in \{1,2,3,4\}$ and setting $\outputguess'_{cont}=\mathbb{I}[\lastlayer'_{cont}>0.5]$ and $\outputguess''_{cont}=\mathbb{I}[\lastlayer''_{cont}>0.5]$ clearly favors the first option over the latter: While we still get $\mutinf{X;\outputguess'_{cont}}=\mutinf{X;\outputguess''_{cont}}=1$ (finite now due to quantization), we have $\mutinf{Y;\outputguess'_{cont}}=\ent{Y}$ but $\mutinf{Y;\outputguess''_{cont}}=0$. 

Including the decision rule, however, does not lead to improved robustness. Indeed, consider Fig.~\ref{fig:robustness}. Then, the resulting RVs $\outputguess^I=\mathbb{I}[\lastlayer^I>0.5]$ and $\outputguess^{II}=\mathbb{I}[\lastlayer^{II}>3/8]$ are identical (and, thus, so are the IB functionals), with the former suffering from reduced robustness. The same can be observed by looking at $\outputguess'_{disc}=\mathbb{I}[\lastlayer'_{disc}>0.5]$ and $\outputguess'''_{disc}=\mathbb{I}[\lastlayer'''_{disc}>0.5]$. Furthermore, this shows that including a decision rule, due to the coarse quantization of $\lastlayer$, leads to large equivalence classes of DNNs that evaluate to the same value in~\eqref{eq:IBOptRule}, which is conceptually similar to the IB functional being piecewise constant (cf.~Sec.~\ref{sec:pwconstant} and~\cite[Sec.~3.5]{Tishby_BlackBox}). 

To apply this method to train intermediate representations other than $L_{m+1}=\lastlayer$, one possible approach is to feed the latent representation $L$ to an auxiliary decision rule $\delta{:}\ \mathbb{R}^{|L|}\to\mathcal{Y}$ and minimizing~\eqref{eq:IBOptRule} for $\outputguess=\delta(L)$. This subsequently leads to layer-wise training in a greedy manner, similar to one discussed in \cite{LayerWise}. Such a layer-wise greedy training procedure must be carefully designed in order to exploit the full benefits of deeper DNN architectures.

\subsection{Probabilistic Interpretation of the Neuron Outputs}\label{sec:probabilistic}
Another option related to Sec.~\ref{subsec:IBsoldec} is to introduce a soft decision rule. For example, in a one-vs-all classification problem with a softmax output layer $\lastlayer$ with $|\mathcal{Y}|$ neurons, the $i$-th entry of $\lastlayer$ can be interpreted as the probability that $\outputguess=i$. Thus, $\outputguess$ is a discrete RV with alphabet $\mathcal{Y}$ that depends stochastically (and not deterministically) on the feature vector $X$. Using this approach not only guarantees that the functional in~\eqref{eq:IBOptRule} is finite but also, unlike in Sec.~\ref{subsec:IBsoldec}, admits applying gradient-based optimization techniques even for finite datasets. Moreover, using a soft decision rule makes the precision term sensitive to simplification, encouraging this property in the output. The precision term also promotes P$2$ in the sense of encouraging $|\lastlayer|$ dense clusters in $\mathbb{R}^{|\lastlayer|}$. These claims can be verified, for example, by looking at $\lastlayer'_{cont}$, $\lastlayer''_{cont}$, and $\lastlayer'''_{cont}$ in Fig.~\ref{fig:problem} (assuming $\lastlayer'_{cont}$, $\lastlayer''_{cont}$, and $\lastlayer'''_{cont}$ are uniform over their support), and identifying values of $\lastlayer$ as probabilities that $\outputguess=1$. The utility of the compression term $\mutinf{X;\outputguess}$ becomes even more questionable than in Sec.~\ref{subsec:IBsoldec}. For one, $\outputguess$ is discrete which automatically enforces implicit compression. Moreover, $\mutinf{X;\outputguess}= \ent{\outputguess} - \ent{\outputguess|X} = \ent{\outputguess} - \ent{\outputguess|\lastlayer}$ is smaller for $\lastlayer=\lastlayer'_{cont}$ than for $\lastlayer=\lastlayer'''_{cont}$ in Fig.~\ref{fig:problem}, hence minimizing $\mutinf{X;\outputguess}$ now prefers $f'_{cont}$ over $f'''_{cont}$ rather than evaluating them equally. Therefore, one either should choose $\beta$ in~\eqref{eq:IBOptRule} more carefully or drop the compression term altogether.

To apply this method to a latent representation, one possible approach is to feed the latent representation $L$ to a linear layer of size $\dom{Y}$ followed by a softmax layer to generate $\outputguess$. Similar to the case in Sec.~\ref{subsec:IBsoldec}, this subsequently leads to layer-wise training in a greedy manner and hence must be be carefully designed in order to exploit the full benefits of deeper DNN architectures.

\subsection{Stochastic DNNs}\label{sec:solstochastic}
A further approach is to use the IB functional to train stochastic DNNs rather than deterministic ones. A DNN can be made stochastic by, for example, introducing noise to the intermediate representation(s). The statistics of the introduced noise can also be considered trainable parameters or adapted to (the statistics of) the intermediate representation(s). The objective function to be optimized remains~\eqref{eq:IBOpt}. For $\mutinf{X;L}$ to be finite, it suffices to add noise with an absolutely continuous distribution to $L$. This approach can be used for layer-wise training as well as for training the DNN as a whole. Depending upon where and what type of noise is introduced, $\mutinf{Y;L}$ can encourage robust representations, for which $\mutinf{Y;L}$ does not degrade by introduction of noise and/or deformations. Similarly $\mutinf{Y;L}$ may also promote intermediate representation with well separated (sub-)regions corresponding to different labels which in turn admit simpler decision functions for the stochastic DNN. For example, a small amount of uniform noise added to the intermediate representation leads to a better IB functional for  $f'_{cont}$ than $f''_{cont}$ and for $f'''_{cont}$ than $f'_{cont}$ in Fig.~\ref{fig:problem}. For stochastic DNNs, the compression term $\mutinf{X;L}$ can encourage more compact representations. For example, again in Fig.~\ref{fig:problem}, adding a small amount of noise $\eta$ to the output makes $\mutinf{X;f'_{cont}(X_1)+\eta}$ larger than $\mutinf{X;f'''_{cont}(X_1)+\eta}$, making the latter representation more desirable. Note that in case of stochastic DNNs, the noisy intermediate representation, such as $f'_{cont}(X_1)+\eta$ or  $f''_{cont}(X_1)+\eta$, is fed as input to the next layer of the DNN.

In addition to resolving the issues associated with IB functional mentioned in Sec.~\ref{sec:IBfails}, training a stochastic DNN in such a way also provides a novel way of data augmentation. Sampling the intermediate representation $L$ multiple times during training for each input sample can be viewed as a way of dataset augmentation, which may lead to improved robustness. Introducing noise in a latent (bottleneck) representation thus presents an alternative to the data augmentation approach proposed in~\cite{DataAugmentation}, which requires training a separate auto-encoder to obtain latent representations to be perturbed by noise.

\subsection{Replacing the IB Functional}\label{sec:solreplace}
A final approach is to replace the IB functional by a cost function that is more well-behaved, but motivated by the IB framework. Specifically, by replacing mutual information by (not necessarily symmetric) quantities $\barmutinfc{\cdot;\cdot}$ and $\barmutinfp{\cdot;\cdot}$, we replace~\eqref{eq:IBOpt} by
\begin{equation}\label{eq:IBOptnew}
 \min_{\{\Theta_i\}} \barmutinfc{X;L} - \beta\barmutinfp{Y;L}.
\end{equation}
This approach can be used  for training the DNN both as a whole and layer-wise.

We first consider setting $\barmutinfc{X;L}=\mutinf{Q_X(X);Q_L(L)}$ and $\barmutinfp{Y;L}=\mutinf{Y;Q_L'(L)}$, where $Q_X$, $Q_L$, and $Q_L'$ are quantizers, that are adapted according to the statistics of the latent representation $L$ and w.r.t. one another\footnotemark. It is important to note that the quantization is not performed inside the DNN, but only for computing the function in~\eqref{eq:IBOptnew}. This is the typical approach performed when mutual information is estimated from finite datasets using histogram-based methods. Unlike \cite{Tishby_BlackBox}, we argue that the design of $Q_X$, $Q_L$, and $Q_L'$ should not only be guided by the goal of estimating the true mutual information (which is bound to fail according to our analysis in Sec.~\ref{sec:infinity}), but also by the aim to instill the desired properties from Sec.~\ref{sec:regularization}  into the cost function \eqref{eq:IBOptnew}.

The effect of quantization is that $\barmutinfc{X;L}$ becomes finite. Moreover, if $Q_{L}'$ is set appropriately, solving~\eqref{eq:IBOptnew} leads to simpler representations. Considering again Fig~\ref{fig:problem},  setting $Q_{L}'(\cdot)=\mathbb{I}[\cdot>0.5]$ prefers $f'_{disc}$ over $f''_{disc}$ (and similarly $f'_{cont}$ over $f''_{cont}$); however, the finer the quantization $Q_L'$ is, the less sensitive is~\eqref{eq:IBOptnew} to the simplicity of the intermediate representation $L$. 

The fact that $Q_L$ and $Q_L'$ need not coincide yields an advantage over the solution proposed in Sec.~\ref{subsec:IBsoldec} in the sense that the compression term can become useful now. With the above choice of $Q_{L}'$ we see that the precision term is the same for $f'_{cont}$ and $f'''_{cont}$. However, if $Q_X$ is the identity function and $Q_{L}$ a uniform quantizer with four quantization levels in $[0,1]$, then $f'_{cont}$ leads to a larger compression term than $f'''_{cont}$, thus favoring $f'''_{cont}$. However, similarly as we observed in Sec.~\ref{subsec:IBsoldec}, the quantized IB functional partitions DNNs into large equivalence classes that do not necessarily distinguish according to robustness. Additionally, the quantized IB functional is piecewise constant when used for finite datasets. Finally, choosing appropriate quantizers is not trivial; e.g., the effect of this choice has been empirically evaluated in~\cite{Anonymous_IBTheory}, with focus only on the compression term. For the quantized IB functional, choosing quantizers becomes even more complicated. 

Without going into details, we note that computing a noisy IB functional, for example by setting $\barmutinfc{X;L}=\mutinf{X;L+\eta}$ and $\barmutinfp{Y;L}=\mutinf{Y;L+\eta'}$ for noise variables $\eta$ and $\eta'$ that are adapted according to (the statistics of) $L$ and w.r.t. each other\footnotemark[\value{footnote}], can lead to simplified and compact representations. In contrast to the quantized IB functional, the noisy IB functional can even lead to robust representations and, for appropriately chosen noise models, is not piecewise constant for finite datasets, hence admitting efficient optimization using gradient-based methods. Again, we note that noise is not introduced inside the DNN, but only in the computation of~\eqref{eq:IBOptnew}; hence, the DNN is still deterministic.

\footnotetext{It is important to adapt the quantizers (or the noise levels) to (the statistics of) the latent representation $L$ in order to rule out ways to decrease the cost without fundamentally changing the characteristics of $L$, e.g., by simple scaling.}

Other than quantization and introducing noise in the computation of the mutual information terms, one may go one step further and replace these terms with different quantities. For example, it is common to replace the precision term by the cross-entropy between the true conditional distribution of $Y$ given $L$ and, e.g., a parametric surrogate distribution. Moreover, also the compression term can be replaced by terms that are inspired by $\mutinf{X;L}$, but differ in essential details. These changes to the optimization problem often directly enforce goals such as P$2$-P$4$, even though they have not been captured by the original optimization problem.

Finally, when replacing the two terms in~\eqref{eq:IBOpt} with different quantities, one may even choose to select different intermediate representations for the precision term and for the compression term. For example, the compression term can be defined based on a latent representation and the precision term can be defined based on $\lastlayer$. The compression term can then enforce desired properties on the latent representation whereas the precision term can ensure that the output of the DNN $\lastlayer$ admits simple decisions and predicts $Y$ well enough. In contrast, evaluating~\eqref{eq:IBOpt} only for an internal representation $L$ trains only the encoder, failing to instill desired properties into $\lastlayer$; evaluating~\eqref{eq:IBOpt} only for the output $\lastlayer$ trains the whole DNN, but does not necessarily lead to internal representations $L$ with the desirable properties.

It is worth mentioning that the approaches in this section are not completely independent. For example, on the one hand, a probabilistic interpretation of the output (Sec.~\ref{sec:probabilistic}) can be considered a special type of stochastic DNN in which the stochasticity appears only in the decision rule. On the other hand, evaluating the IB functional for this probabilistic interpretation can be considered as replacing the IB functional with a different cost function. This is in line with the reasoning in~\cite{Achille_InfoDropout}, illustrating that the same problem may be solved equivalently by adapting the optimization method, the feasible set, or the cost function.

A common theme in Secs.~\ref{subsec:IBsoldec} to~\ref{sec:solreplace} is that our remedies encourage latent representations in which data points from different classes are represented in some geometrically compact manner. In other words, the proposed remedies encourage compression (P$2$) in a geometric sense rather than in the sense of a minimal sufficient statistic. This is intuitive, since representing classes by clusters tight and far apart allows using simple decision rules for classification (P$3$). While such clustering does not immediately ensure robustness (P$4$), the injection of noise, either directly (Sec.~\ref{sec:solstochastic}) or only in the computation of the IB functional (Sec.~\ref{sec:solreplace}) does.

All this certainly does not imply that measuring P$2$ in information-theoretic terms is inadequate. Rather, it illustrates that measuring P$2$ in information-theoretic terms is insufficient to instill desirable properties such as simple decision functions or robustness, while understanding P$2$ in geometric terms has the potential to do so.

\section{Critical Discussion of and Experimental Evidence from Related Work}\label{sec:related}
The idea of using the IB framework for DNNs was first introduced in \cite{Tishby_DLIB_ITW}. They proposed using the IB functional to analyze DNNs in terms of performance as well as architectural compactness and argued that this can be done not only for the output but also for hidden layer representations. This, purportedly, leads to a deeper insight into the inner workings of a DNN than an evaluation based on the output or network parameters could. They also suggested the IB functional as an optimization criterion for training DNNs.

Following~\cite{Tishby_DLIB_ITW}, several works have been published that are based on the application of the IB framework to DNNs and the contribution of which can be loosely grouped into the following three categories: a) experimental IB-based DNN analysis, b) IB-based DNN performance bounds and IB theory of DNNs, and c) IB-based DNN training. We now briefly review the most prominent works in either category and connect this analysis to our work.

The most prominent work in the first category is~\cite{Tishby_BlackBox}, in which the authors applied these ideas to analyze DNNs trained using cross-entropy without regularization. They empirically observed that the individual layers of a DNN lead to representations that are close to optimal in the sense of the IB functional for a given trade-off parameter $\beta$. They also observed that compression cannot be linked to architectural simplicity but claimed, based on experiments, that stochastic gradient descent exhibits a compression phase that, they believe, is causally linked to improved generalization performance.

Several authors have attacked~\cite{Tishby_BlackBox} since, which initiated a debate that is still ongoing. Specifically, the authors of~\cite{Anonymous_IBTheory} discussed analytically and empirically that the compression phase observed in~\cite{Tishby_BlackBox} is an artifact of the quantization strategy used to approximate the compression term in connection with the activation function used. They furthermore present experimental evidence suggesting that compression and generalization are phenomena that may or may not occur simultaneously. It has to be noted, though, that both~\cite{Tishby_BlackBox} and~\cite{Anonymous_IBTheory} quantify compression not via $\mutinf{X;L}$ (which we have shown to be infinite in Sec.~\ref{sec:infinity}), but via mutual information estimators (e.g., binning estimators).  We believe that our Sec.~\ref{sec:IBfails} provides an interesting perspective on~\cite{Tishby_BlackBox,Anonymous_IBTheory}: First, the computability/optimizability issues in Sec.~\ref{sec:infinity} and Sec.~\ref{sec:pwconstant} imply that $\mutinf{X;L}$ can be used to study the training dynamics of a deterministic DNN only in severely restricted scenarios. Second, the observations in Sec.~\ref{sec:invariance} indicating that the IB functional fails to capture desired properties casts serious doubts about the possibility of using $\mutinf{X;L}$, without modification, to make claims about generalization performance of a deterministic DNN. Besides, recently proposed invertible DNN architectures~\cite{iRevnet, reversiblenets} are shown to achieve state-of-the-art performance despite the fact that they do not compress at all in an information-theoretic sense. Indeed, due to their invertibility, $\mutinf{X;L_m}$ is the same for every layer regardless of the network parameters. This suggests that information-theoretic compression, if at all, cannot be the only cause of good generalization performance.

The utility of a compression term is partly rehabilitated in~\cite{zivinfo}, which studied the latent representations obtained via training a DNN using a standard loss function. They inject Gaussian noise at the output of every neuron and show that, in this case, a geometric clustering of the latent representation is captured well by both the compression term $\mutinf{X;L}$ and the entropy of the quantized latent representation, $\ent{Q_L(L)}$. As we mention at the end of Sec.~\ref{sec:hotwo}, encouraging geometric clustering has the potential to directly instill the desirable properties of simple decision rules (P$3$) and robustness (P$4$) into latent representations. The observations in~\cite{zivinfo} therefore support our proposal to either use a stochastic DNN (Sec.~\ref{sec:solstochastic}) or to replace the cost function, e.g., via quantized entropy (Sec.~\ref{sec:solreplace}).

The second of the three categories is mainly populated by analytical results. For example,~\cite[Sec.~2~\&~Appendix~C]{Anonymous_IBTheory} briefly investigated the computability issues of the compression term $\mutinf{X;L}$ in the IB functional and recognized that this term is infinite if the intermediate representation is continuous, a result which we generalize in Theorem~\ref{thm:infinity} in Sec.~\ref{sec:infinity}. To remedy this issue, the authors suggest to replace the compression term by the mutual information between $X$ and a noisy or quantized version of $L$ (cf.~Sec.~\ref{sec:solreplace}). For discrete $X$ and $L$, the authors of~\cite{ISITIBtheory} use the finite quantity $\mutinf{X;L}$ to bound the generalization gap from above. Although theoretically interesting, this bound relies on strong assumptions such as the use of an \enquote{optimal} decoder for the latent representation $L$. Furthermore, the upper bound only accounts for the generalization gap and not the actual performance (which can be poor despite a small generalization gap). The upper bound is also infinite, and hence not useful, in the setting of deterministic DNNs with continuous X and L (see Sec.~\ref{sec:infinity}). Finally, Kolchinsky et al.\ investigated the IB functional for the special case in which the class $Y$ is a deterministic function of the features $X$~\cite{Kolchinsky_Pathologies}. In this case, the individual layers of a DNN can only explore a weak trade-off between compression and prediction, as the prediction cannot decrease towards deeper layers (although compression may still increase). This is not in conflict with the claims of~\cite{Tishby_BlackBox}, but complements them in an interesting special case. This observation also agrees with the desirable properties P$1$ and P$2$ that we introduced in Sec.~\ref{sec:regularization}, as deeper representations can be compressed more (due to a more complex encoder) while still being a sufficient statistic for $Y$.

We finally turn to the third category comprised of works that train DNNs using cost functions inspired by the IB principle. As we explain in the discussion that follows, these works employ different approaches proposed in Sec.~\ref{sec:hotwo}. The successes reported in some of these works are in terms of various operational goals that are directly relevant for applying classifiers in practice, such as generalization, adversarial robustness, and out-of-distribution detection. Based on our discussion in Sec.~\ref{sec:IBfails}, however, these results cannot be attributed to the use of IB functional and should, therefore, at least partly be considered an outcome of these additional modifications from Sec.~\ref{sec:hotwo}. Hence the works discussed in this category provide experimental evidence that our proposed remedies are not only successful in instilling desired characteristics in intermediate representations, such as P$3$ and P$4$, but are also, via these characteristics of the intermediate representations, able to successfully achieve various operational goals.

The authors of~\cite{Achille_InfoDropout} proposed minimizing the IB functional based on parametric distributions for the (stochastic) encoder and decoder, i.e., combining approaches from Secs.~\ref{sec:solstochastic} and~\ref{sec:solreplace}. They showed that minimizing the cost function (regularized by total correlation to encourage disentangled representations) is equivalent to minimizing cross-entropy over DNNs with multiplicative noise (dubbed \emph{information dropout}). They also discovered that, for a certain choice of the parameter $\beta$ and for the goal of reconstruction, i.e., $Y\equiv X$, the regularized cost function is equivalent to the one for variational auto-encoding. 

As a second prominent example, the authors of~\cite{Alemi_DVIB} trained a stochastic DNN using a variational upper bound on IB functional and showed that the resulting DNN has state-of-the-art generalization performance as well as improved robustness to adversarial examples. They introduce noise at a dedicated bottleneck layer (cf.~Sec.~\ref{sec:solstochastic}), leading to a stochastic DNN with finite IB functional for the bottleneck and the subsequent layers. The authors then replace the compression term $\mutinf{X;L}$ for the bottleneck layer with a variational upper bound (cf.~Sec.~\ref{sec:solreplace}) to make the compression term tractable; the resulting term is no longer invariant under bijections and encourages bottleneck representations that are compact in a geometric sense. They further replace the precision term $\mutinf{Y;L}$ by cross-entropy loss (cf.~Sec.~\ref{sec:solreplace}. In Sec.~\ref{sec:precisionapprox} we interpret this as two steps applied sequentially, namely first lower bounding $\mutinf{Y;L}$ by $\mutinf{Y;\lastlayer}$ and then lower bounding $\mutinf{Y;\lastlayer}$ by cross-entropy loss with a probabilistic interpretation of the output $\lastlayer$ (cf.~Sec.~\ref{sec:probabilistic}). Combining the bounds on compression and precision terms thus instill desirable properties in both the bottleneck representation $L$ and the output representation $\lastlayer$ (cf.~end of Sec.~\ref{sec:solreplace}). Unlike $\mutinf{Y;L}$ and $\mutinf{Y;\lastlayer}$, cross-entropy applied to a probabilistic interpretation of the output is no longer insensitive to bijections and, in conjunction with noise introduced at the bottleneck layer, enforces simplicity of the decision rule (P$3$) and robustness (P$4$) in the trained DNN (see~\cite[Fig.~2]{Alemi_DVIB}).

Several further works suggest IB-based training methods similar in spirit to~\cite{Alemi_DVIB}. The authors of~\cite{Kolchinsky_NLIB} optimized the IB functional using stochastic DNNs, which is closely related to training stochastic DNNs using the IB functional. Rather than replacing the compression term by a variational bound as in~\cite{Alemi_DVIB}, the propose a non-parametric bound. As a result, the intermediate representations they obtain form geometrically dense clusters (see ~\cite[Fig.~2]{Kolchinsky_NLIB}). The authors of~\cite{uncertaintyDVIB} used the same technique as in~\cite{Alemi_DVIB} to train stochastic neural networks but they measure the performance of the DNNs in terms of classification calibration as well the DNN's ability to detect out-of-distribution data. Finally, the authors of~\cite{vdb} introduce Gaussian noise at the latent representation (Sec.~\ref{sec:solstochastic}) and approximate $\mutinf{X;L}$ similar as~\cite{Alemi_DVIB} does. However, they replace the precision term $\mutinf{Y;L}$ in the IB functional with a term that quantifies channel deficiency based on Kullback-Leibler divergence and use a tractable approximation for this new precision term (Sec~\ref{sec:solreplace}). Note that the authors of \cite{Alemi_DVIB,Kolchinsky_NLIB,uncertaintyDVIB,vdb} do not explicitly mention that the IB functional leads to an ill-posed optimization problem the solution of which lacks desirable properties such as representational simplicity and robustness, rather they introduced the aforementioned modifications to obtain tractable bounds on the IB functional that are optimizable using gradient based methods.

In summary, the majority of works published agree, either explicitly or implicitly that, without modification, the IB framework is neither fit for analyzing nor for training DNNs: For analysis, the compression term 
$\mutinf{X;L}$ was discovered to be infinite (compare~\cite{Anonymous_IBTheory,zivinfo} with our Sec.~\ref{sec:infinity}) or irrelevant (compare~\cite{iRevnet, reversiblenets} with our Secs.~\ref{subsec:IBsoldec} and~\ref{sec:probabilistic}) and is best replaced by a term capturing ``geometric'' compression (compare~\cite{zivinfo,Alemi_DVIB,Kolchinsky_NLIB} with the end of our Sec.~\ref{sec:hotwo}). For training, the authors of~\cite{Achille_InfoDropout,Alemi_DVIB,Kolchinsky_NLIB,uncertaintyDVIB,vdb} propose cost functions motivated by the IB framework but, for various reasons, depart from it by using a combination of techniques from Sec.~\ref{sec:hotwo}. This not only provides evidence that our analysis in Sec.~\ref{sec:IBfails} is relevant and valid, but also that the promising performance of~\cite{Achille_InfoDropout,Alemi_DVIB,Kolchinsky_NLIB,uncertaintyDVIB,vdb} in terms of various operational goals can at least partly be attributed to the remedies we propose in Sec.~\ref{sec:hotwo}.

\section{Concluding Remarks}\label{sec:discussion}
We have shown in Sec.~\ref{sec:IBfails} that training deterministic DNNs using the IB functional suffers from serious problems. Aside from the optimization problem~\eqref{eq:IBOpt} being ill-posed (Sec.~\ref{sec:infinity}) or inaccessible to gradient-based optimization (Sec.~\ref{sec:pwconstant}), the IB functional does not capture desirable properties of intermediate representations, such as allowing simple decisions and robustness to noise (Sec.~\ref{sec:invariance}). Including a simple decision rule while computing the IB functional solves some of these problems, but may lead to cost functions that are piecewise constant (Sec.~\ref{subsec:IBsoldec}). Similarly, training stochastic DNNs with the IB functional (Sec.~\ref{sec:solstochastic}) solves some problems and additionally provides a new mechanism for task-specific data augmentation. Most obviously, replacing the IB functional with a cost function that is more well-behaved also leads to robust and simple representations (Sec.~\ref{sec:solreplace}). To achieve this goal it is not even necessary to depart far from the IB framework: Replacing $\mutinf{Y;\lastlayer}$ by $\mutinf{Y;\lastlayer+\eta'}$ and $\mutinf{X;\lastlayer}$ by $\mutinf{X;\lastlayer+\eta}$, for noise terms $\eta'$ and $\eta$ can lead to DNNs that are robust and the output of which allows using a simple decision rule. Sec.~\ref{sec:related} not only critically assesses the related work but also utilizes it to provide empirical evidence for the success of the remedies we propose in Sec.~\ref{sec:hotwo}.

We wish to mention that the discussion in Sec.~\ref{sec:invariance} holds equally well for the analysis of DNNs using the IB framework: A good result in terms of the IB functional does neither admit statements about the robustness of the DNN nor about the simplicity of the required decision function. The ill-posed or piecewise constant nature of IB functional for the classification task using DNNs (cf. Sec.~\ref{sec:infinity} and Sec.~\ref{sec:pwconstant}) further complicates the situation and makes it an unfit tool for analysis.  Our results regarding generalization are thus in line with the observations in~\cite{Anonymous_IBTheory}. 

We believe that the idea of regularization introduced by the IB functional, i.e., to regularize the intermediate representations $L$ rather than the parameters  $\{\Theta_i\}$ of the DNN, has great potential. Traditional complexity measures focus on what a DNN can do based on the network architecture while ignoring the (estimated) data statistics $P_{X,Y}$ and the actual function implemented by the DNN, i.e., the network parameters after training on the actual data. For example, VC dimension is independent of $P_{X,Y}$ and the learned network parameters, and the generalization error bound based on Rademacher complexity only depends on (estimated) $P_X$, ignoring $P_{Y|X}$ and the learned network parameters. It has been noted that these traditional measures fail to explain why largely over-parameterized DNNs generalize well although they are capable of memorizing the whole dataset~\cite{novaksensitivity,rethinkinggeneralization, DisentangledRepresentations}.  The discussion in \cite{memorizationDNN} suggests  that a perspective of understanding the DNN capacity, which also involves the (estimated) relation between $X$ and $Y$ as well as the learned network parameters, can lead to more meaningful insights; the methods employed in \cite{novaksensitivity,zahavyrobust, iRevnet} for explaining the success of such networks explicitly or implicitly focus on the learned representations for the specific task. 

All this hints at the fact that regularization based on properties of intermediate representations can be beneficial. The experiments in~\cite{zivinfo} also support this claim by arguing that, e.g., geometric clustering of the latent representations is a valid goal for training DNNs for classification. Such a direct design can also ensure compatibility with standard optimization tools used in deep learning, such as gradient-based training methods.

In the recent literature, several regularizers trying to instill desired characteristics directly in an intermediate representation (without necessarily being motivated by the IB principle) have been proposed.
Reference~\cite{invariantDVIB} introduces $\mutinf{C;L}$, where $C$ is some nuisance factor or discriminatory trait, as a regularizer. Minimizing it makes the latent representation $L$ and the performance of the DNN invariant to $C$. The authors evaluate this regularization in a variational auto-encoding setup and as an additional term in the variational IB setup of~\cite{Alemi_DVIB}. Depending upon the relation between $X$ and $C$, the term $\mutinf{C;L}$ may also suffer from the issues discussed in Sec.~\ref{sec:infinity} or Sec.~\ref{sec:pwconstant} for deterministic DNNs.
In~\cite{OutputRegularizer}, the authors regularized the final softmax output, interpreted as a probability distribution over $\mathcal{Y}$ (cf. $\outputguess$ in Sec.~\ref{sec:probabilistic}), by  $\ent{\outputguess|X}$ to penalize overly confident output estimates. The experiments in~\cite{latentreg1} suggest that back-propagation of classification error or a similar loss function from the output does not lead to latent representations with properties such as discrimination and invariance (which are intuitively similar to P$1$ and P$3$ and to P$2$ and P$4$ from Sec.~\ref{sec:regularization}, respectively).
They propose a ``hint penalty'' that encourages latent representations being similar if they correspond to the same class. Similarly,~\cite{latentreg2} defined the regularization on latent representations as a label-based clustering objective, which is conceptually similar to defining the goal of compression (P$2$) in a geometric sense. The authors of~\cite{latentreg2} discuss the performance of such regularizers for various problems including auto-encoder design, classification, and zero shot learning.

In addition to improved generalization, carefully designing representation-based regularizers may have additional advantages. Enhanced adversarial robustness for such networks has been demonstrated in \cite{Alemi_DVIB}. Such regularization also provides a more flexible data augmentation method for training and inference as compared to the fixed transformations done at the input currently in practice, e.g. rotation, translation etc. For example, the authors of~\cite{Alemi_DVIB,Kolchinsky_NLIB} sample the noisy bottleneck representation $L$ multiple times for each input training example. This data augmentation mechanism also yields an advantage over the one introduced in \cite{DataAugmentation} by obviating the need to train a separate auto-encoder and gets automatically adapted to the classification task at hand. Regularizers can also be used to enforce privacy guarantees or to ensure insensitivity to transformations such as rotations, translations, etc.\ which is attributed to be one cause of the superior performance of DNNs \cite{DisentangledRepresentations}. 

All of these works and those discussed in Sec.~\ref{sec:related} provide empirical evidence that  regularizing latent representation(s) is a promising endeavor, achieving generalization, robustness, fairness, classification calibration, and data augmentation. On the one hand, the discussion at the end of Sec.~\ref{sec:hotwo} and in \cite{zivinfo} along with the empirical evidence from \cite{latentreg1,latentreg2,iRevnet} leads us to believe that defining  a latent representation regularizer in a geometric sense in conjunction with noise/stochasticity is a promising domain of future research. On the other hand, although regularizing latent representations is a key feature of the IB framework, it fails to instill desired properties in the latent representations. The success of invertible DNNs, e.g., iRevnet \cite{iRevnet}, and our analyses suggest that the information-theoretic compression-based regularization term $\mutinf{X;L}$ either becomes obsolete or has to be replaced. Similarly, although intuitively often attractive, if one aims to define the latent regularizer via some other information-theoretic cost (e.g., as in~\cite{invariantDVIB}), it is important to mitigate issues including, but not limited to, the ones discussed in Sec.~\ref{sec:infinity}, Sec.~\ref{sec:pwconstant}, and Sec.~\ref{sec:invariance}. In contrast, it can be observed by looking at, e.g., \cite{Kolchinsky_NLIB, Alemi_DVIB} that a restriction to a specific prior distribution and approximations being used to evaluate the information-theoretic cost lead to a more direct and intuitive geometric interpretation which can be utilized. Thus, in light of the discussion in this work and in \cite{novaksensitivity,zivinfo,iRevnet} along with the empirical evidence in \cite{latentreg1,latentreg2}, we conclude that designing regularizers directly with the aim of instilling certain properties desirable for the intermediate representation $L$ (such as discussed in Sec.~\ref{sec:regularization}) may be a more fruitful approach than trying to repair the problems inherent in the IB functional (or other information-theoretic cost functions) in the context of classification.

\section{Proof of Theorem~\ref{thm:infinity}}\label{sec:proof}

We denote vectors by lower-case letters, i.e., we write $z\triangleq(z_1,\dots,z_N)\in\mathbb{Z}^N$. Moreover, we define the $N$-dimensional cube with side length $a$ and bottom-left corner at $z\in\reals^N$ as $\cube{z}{a}\triangleq[z_1,z_1+a)\times\cdots[z_N,z_N+a)$. For example, the RV $\qRVm{X}\triangleq\lfloor mX\rfloor/m$, where the floor operation is applied element-wise, is obtained by quantizing $X$ with a quantizer that has quantization bins $\cube{z/m}{1/m}$, $z\in\mathbb{Z}^N$.

Let $\ent{Z}$ denote the entropy of the discrete RV $Z$ with probability mass function $\pmf{Z}$ and alphabet $\dom{Z}$, and let
\begin{equation}\label{eq:ent2}
 \enttwo{Z} \triangleq -\log \left(\sum_{z\in\dom{Z}} \left(\pmf{Z}(z)\right)^2 \right).
\end{equation}
denote the R\'enyi entropy of second order of $Z$. The correlation dimension of a general RV $X$ is defined as~\cite{Csiszar_InfoDim} 
\begin{equation}\label{eq:cordim}
 \cordim{X}\triangleq\lim_{m\to\infty} \frac{\enttwo{\qRVm{X}}}{\log m}
\end{equation}
provided the limit exists. The information dimension $\infodim{X}$ is defined accordingly, with R\'enyi entropy of second order replaced by entropy~\cite{Renyi_InfoDim}.

\begin{proof}[Proof of the Theorem]
The proof consists of four ingredients. Assuming that the distribution of $X$ has a continuous PDF supported on a compact set, we first show that the input $X$ has positive correlation dimension. Then, we show that the correlation dimension remains positive throughout the DNN. Afterwards, we show that the output has positive information dimension, from which follows that $\mutinf{X;L}=\infty$. Finally, we relax the condition that $X$ has a continuous PDF supported on a compact set, but require that its distribution has at least such a component.

We start by assuming that $X$ has a continuous PDF that is supported on a compact set $\mathcal{X}$ in $\reals^N$.

\begin{lem}\label{lem:cordim}
 Let $X$ be an $N$-dimensional RV with a PDF $f_X$ that is continuous and supported on a compact set $\mathcal{X}$ in $\reals^N$. Then, $\cordim{X}=N$.
\end{lem}

This result generalizes~\cite[Th.~3.I.c]{Csiszar_InfoDim} to higher-dimensional RVs.

\begin{proof}
 Since $f_X$ is continuous, so is its square $f_X^2$. Since both $f_X$ and $f_X^2$ are continuous and supported on a compact set, they are Riemann integrable. Hence, the differential R\'enyi entropy of second order,
\begin{equation}
 h^2(X) \triangleq - \log \int_\mathcal{X} f_X^2(x) \mathrm{d}x
\end{equation}
exists. We can sandwich $h^2(X)$ by using the upper and lower Darboux sums, i.e., we can write
\begin{align*}
  &L_{h^2,m}\notag\\&\triangleq -\log \left(\sum_{z\in\mathbb{Z}^N} \frac{1}{m^N} \sup_{x\in \cube{z/m}{1/m}} f_X^2(x) \right)\\
 &\le h^2(X)\\
&\le -\log \left(\sum_{z\in\mathbb{Z}^N} \frac{1}{m^N} \inf_{x\in \cube{z/m}{1/m}} f_X^2(x) \right) \triangleq U_{h^2,m}.
\end{align*}
Note further that by the mean value theorem we can find $t_z\in \cube{z/m}{1/m}$ such that 
\begin{equation}
 \pmf{\qRVm{X}}(z/m)= \int_{x\in \cube{z/m}{1/m}} f_X(x)\mathrm{d}x = \frac{1}{m^N} f_X(t_z).
\end{equation}
This allows us to write $  \enttwo{\qRVm{X}}$ as
\begin{equation}
 \enttwo{\qRVm{X}} = -\log \left( \sum_{z\in\mathbb{Z}^N} \frac{1}{m^{2N}} f_X^2(t_z) \right)
\end{equation}
Since $f_X^2(t_z)$ lies between the infimum and the supremum $f_X$ can assume on the cube $\cube{z/m}{1/m}$, we obtain
\begin{equation}
 L_{h^2,m}\le \enttwo{\qRVm{X}} - N\log m \le U_{h^2,m}
\end{equation}
where, because of Riemann integrability, the outer terms of this inequality become equal as $m$ tends to infinity. Hence, with~\eqref{eq:cordim}, $\cordim{X}=N$.
\end{proof}

Now suppose that $L_i$ has a distribution with correlation dimension $\cordim{L_i}$ and compact support $\dom{L}_i\subset\reals^{|L_i|}$. Then, we have by~\cite[Th.~1.1]{Hunt_Projections}
\begin{equation}\label{eq:projPreservation}
 \cordim{\mathbf{W}_i^T L_i} = \min\{|L_i|,|L_{i+1}|,\cordim{L_i}\}
\end{equation}
for almost every $|L_i|\times |L_{i+1}|$ matrix $\mathbf{W}_i$ (in the sense of the Lebesgue measure on the space of $|L_i|\times |L_{i+1}|$ matrices). Since $\dom{L}_i$ is compact, so is the support $\dom{L}_i'$ of the distribution of $\mathbf{W}_i^T L_i+b_{i+1}$. Thus, we can find a rectangle $\dom{L}''_i\triangleq\dom{L}''_{i,1}\times\cdots\times\dom{L}''_{i,|L_{i+1}|}\subset\reals^{|L_{i+1}|}$ that contains $\dom{L}_i'$. Suppose the activation function $\sigma$ is continuously differentiable with strictly positive derivative. Thus, it is strictly monotonically increasing and has a strictly monotonic inverse that, by the inverse function theorem, is continuously differentiable. Let $\dom{L}''_{i+1,j}$ be the image of $\dom{L}''_{i,j}$ under $\sigma$; since $\dom{L}''_{i,j}$ is compact, so is $\dom{L}''_{i+1,j}$. It follows that the function $\sigma{:}\ \dom{L}''_{i,j}\to\dom{L}''_{i+1,j}$ is bi-Lipschitz, hence so is the function mapping $\mathbf{W}_i^T L_i+b_{i+1}$ to $L_{i+1}$. Since bi-Lipschitz mappings do not affect correlation dimension (see~\cite[Th.~2.6]{Mattila_DimensionOfAMeasure}), we have with~\eqref{eq:projPreservation}
\begin{equation}
 \cordim{L_{i+1}}= \min\{|L_i|,|L_{i+1}|,\cordim{L_i}\}.
\end{equation}
Furthermore, if the support of the distribution of $L_i$ is compact, so is the support of the distribution of $L_{i+1}$. We can thus apply this argument recursively: Indeed, since the distribution of $L_0=X$ has a compact support on $\reals^{N}$, we get with Lemma~\ref{lem:cordim},
\begin{equation}
 \cordim{L_{i+1}} = \min_{j=0,\dots,i+1} |L_j|.
\end{equation}

Correlation dimension bounds information dimension from below (see also~\cite[p.~257]{Wu_DoF}). Hence,
\begin{equation}\label{eq:lowerboundLm}
 \infodim{L_{i+1}} \ge \min_{j=0,\dots,i+1} |L_j|
\end{equation}
i.e., the information dimension of the output of the DNN is positive.

We finally relax the condition that $X$ has a continuous PDF supported on a compact set. By assumption, the distribution of $X$ has an absolutely continuous component that has a PDF $f_X$ that is continuous on a compact set $\dom{X}\subset\reals^N$. The distribution of $X$ therefore is a mixture of an absolutely continuous distribution supported on $\dom{X}$ and some other distribution that may have non-compact support, be discrete, or even singular w.r.t.\ the $N$-dimensional Lebesgue measure. Let $X_c$ denote the RV distributed according to the absolutely continuous component, and let $X_s$ denote the RV distributed according to the remaining component. Let further $Y$ be a binary RV controlling this mixture. In other words, we have $X=X_c$ if $Y=0$ and $X=X_s$ if $Y=1$. Then, the information dimension of $L_{i+1}$ satisfies~\cite[Th.~2]{Wu_Renyi}
\begin{multline}
 \infodim{L_{i+1}} = \mathbb{P}(Y=0)\infodim{L_{i+1}|Y=0}\\ +  \mathbb{P}(Y=1)\infodim{L_{i+1}|Y=1}
\end{multline}
where $\mathbb{P}(Y=0)>0$ by assumption. Since $X=X_c$ if $Y=0$, we can use~\eqref{eq:lowerboundLm} to show that $\infodim{L_{i+1}|Y=0}>0$. Combining this with $\mathbb{P}(Y=0)>0$ yields that $\infodim{L_{i+1}}>0$ regardless of the distribution of $X_s$.

In contrast, since the DNN is deterministic, the conditional distribution of $L_{i+1}$ given $X$ is discrete. It follows that $\infodim{L_{i+1}|X}=0$. Therefore,
\begin{equation}
 \infodim{L_{i+1}} > \infodim{L_{i+1}|X}
\end{equation}
from which we obtain $\mutinf{X;L_{i+1}}=\infty$ with~\cite[Prop.~4.2]{Geiger_LossSystems}. This completes the proof.
\end{proof}

\section{Our Perspective on Bounding $\mutinf{Y;L}$}\label{sec:precisionapprox}
In this section we present an alternative view of the cross-entropy lower bound employed in \cite{Alemi_DVIB,Kolchinsky_NLIB} for the precision term $\mutinf{Y;L}$. We start by restricting ourselves to the setup where, for a latent layer $L$, the encoder $L = f(X)$ can be a stochastic map of $X$ but the decoder $\lastlayer = h(L)$ is a deterministic map. Note that the scenarios in  \cite{Alemi_DVIB,Kolchinsky_NLIB} are subsumed in this setup.

The prevalent perspective, also described in \cite{Kolchinsky_NLIB} and \cite{Alemi_DVIB}, of defining the lower bound is:
\begin{align}\label{eq:precisionlowerbound}
\mutinf{Y;L}  &\geq \ent{Y} - \ent{Y|L} - \kl{P_{Y|L}}{Q_{Y|L}}  \\
&= \ent{Y}  - \cent{P_{Y|L}}{Q_{Y|L}}  \label{eq:boundcross}
\end{align}
where $\kl{\cdot}{\cdot}$ and $\cent{\cdot}{\cdot}$ denote Kullback-Leibler divergence and cross-entropy, respectively, and where $Q_{Y|L}$ is any conditional probability distribution over $\mathcal{Y}$ given $L$ (normally referred to as the variational approximation of the true posterior $P_{Y|L}$).

For deterministic $h$, we make the assumption that for all $y \in \mathcal{Y}$ and $\ell \in \reals^{|L|}$
 \begin{equation}\label{eq:equalvarapprox}
 Q_{Y|L}(y|\ell) = Q_{Y|\lastlayer} (y|h(\ell))
 \end{equation}
 i.e., that $Q_{Y|L}$ depends on $L$ only via the deterministic decoder $\lastlayer =h(L)$. This assumption defines the role that the deterministic decoding map $h$ plays in the variational approximation and holds true for the scenarios described in \cite{Kolchinsky_NLIB, Alemi_DVIB}. $Q_{Y|\lastlayer}$ is again a variational approximation of the true posterior $P_{Y|\lastlayer}$ and it is normally obtained via the application of a simple decision rule to $\lastlayer$. For example, in~\cite{Alemi_DVIB,Kolchinsky_NLIB} the output $\lastlayer$ of the last softmax layer is interpreted as a probability distribution $Q_{Y|\lastlayer}$ over $\dom{Y}$ (as in Sec.~\ref{sec:probabilistic}).

\newcommand{\expecwrt}[2]{\mathbb{E}_{#1}\left(#2\right)}

On the one hand, for $Q_{Y|L}$ satisfying~\eqref{eq:equalvarapprox}, we have
\begin{align}
&\cent{P_{Y|L}}{Q_{Y|L}} =  -\expecwrt{L,Y\sim P_{L,Y} }{\log Q_{Y|L}(Y|L)}  \notag\\
&\stackrel{(a)}{=}  -\expecwrt{L,Y\sim P_{L,Y} }{\log Q_{Y|\lastlayer}(Y|h(L))} \notag\\
&\stackrel{(b)}{=} -\expecwrt{\lastlayer,Y\sim P_{\lastlayer,Y}}{\expecwrt{L\sim P_{L|\lastlayer}(\cdot|\lastlayer)}{\log Q_{Y|\lastlayer}(Y|h(L))}} \notag\\
&\stackrel{(c)}{=}  -\expecwrt{\lastlayer,Y\sim P_{\lastlayer,Y}}{\log Q_{Y|\lastlayer}(Y|\lastlayer)} \notag\\
                                             &= \cent{P_{Y|\lastlayer}}{Q_{Y|\lastlayer}} \label{eq:crossequal}
\end{align}
where $(a)$ is due to~\eqref{eq:equalvarapprox}, $(b)$ due to the law of total expectation, and $(c)$ because $\lastlayer=h(L)$ and, thus, the inner expectation has a constant argument.

On the other hand, we have
\begin{align}
\mutinf{Y;L} &=\mutinf{Y;\lastlayer} + \mutinf{Y;L|\lastlayer} \notag\\
&\stackrel{(a)}{\geq} \mutinf{Y;\lastlayer} \notag\\
&\stackrel{(b)}{\geq} \ent{Y} - \cent{P_{Y|\lastlayer}}{Q_{Y|\lastlayer}}\notag \\
&\stackrel{(c)}{=}\ent{Y} - \cent{P_{Y|L}}{Q_{Y|L}} \label{eq:list_MI}
\end{align}
where $(a)$ follows from the non-negativity of mutual information, $(b)$ is obtained by employing the same technique as in \eqref{eq:boundcross} and $(c)$ is due to \eqref{eq:crossequal}. Inequalities $(a)$ and $(b)$ signify our two-step perspective of lower bounding $\mutinf{Y;L}$ using the cross-entropy loss $\cent{P_{Y|L}}{Q_{Y|L}} $. Specifically, $(a)$ defines the relevant information loss \cite[Chap.~5]{Geiger_LossSystems} due to the transformation of $L$ to $\lastlayer$ via the deterministic decoding map $h$, while $(b)$ represents the loss due the variational approximation $Q_{Y|L} = Q_{Y|\lastlayer}$ obtained by applying the decision rule to $\lastlayer$. Hence, we have divided the total approximation loss into two parts, one corresponding to the decoding map $h$ and the other corresponding to the decision rule, in the same way in which $Q_{Y|L}$ is defined based on $L$ in two steps via $\lastlayer$. This also shows that the cross-entropy loss used in \cite{Alemi_DVIB, Kolchinsky_NLIB} is a better surrogate for $\mutinf{Y;\lastlayer}$ than for $\mutinf{Y;L}$.

Let us now briefly consider a stochastic decoding map $h$ and define
\begin{equation}\label{eq:average}
  Q_{Y|L} (y|\ell) = \expecwrt{\lastlayer\sim P_{\lastlayer|L}(\cdot|\ell)}{Q_{Y|\lastlayer}(y|\lastlayer)}
\end{equation}
which simplifies to~\eqref{eq:equalvarapprox} in case $h$ is deterministic. For stochastic $h$ and $ Q_{Y|L} (y|\ell) $ defined in \eqref{eq:average} we have
\begin{align}\label{eq:crossentropy}
 &\cent{P_{Y|L}}{Q_{Y|L}}  \notag \\
 &= -\expecwrt{L,Y\sim P_{L,Y} }{\log \left(\expecwrt{\lastlayer\sim P_{\lastlayer|L}(\cdot|L)}{Q_{Y|\lastlayer}(Y|\lastlayer)}\right)} \notag\\
  &\stackrel{(a)}{\le} -\expecwrt{L,Y,\lastlayer\sim P_{L,Y,\lastlayer} }{\log Q_{Y|\lastlayer}(Y|\lastlayer)} \notag\\
 &\stackrel{(b)}{=} \cent{P_{Y|\lastlayer}}{Q_{Y|\lastlayer}}.
\end{align}
where $(a)$ follows from Jensen's inequality and $(b)$ is due to law of total expectation. This suggests that for stochastic encoders, $\cent{P_{Y|L}}{Q_{Y|L}}$ provides a tighter lower bound for $\mutinf{Y;L}$ than $\cent{P_{Y|\lastlayer}}{Q_{Y|\lastlayer}}$.

The two cross-entropy terms in~\eqref{eq:crossentropy} also have different operational interpretations in the case of a stochastic decoding map $h$. To understand this better, let us investigate how they are estimated in practice. For each available data sample $(x,y)$, we generate multiple samples of $L$ according to the stochastic map $f=P_{L|X}$. Then, for every $\ell$, we generate multiple samples of $\lastlayer$ according to the stochastic map $h=P_{\lastlayer|L}$. On the one hand, to compute $\cent{P_{Y|L}}{Q_{Y|L}}$, we average $Q_{Y|\lastlayer}$ corresponding to different samples of $\lastlayer$ generated for the same tuple $(x,\ell,y)$ to calculate $Q_{Y|L}$ (i.e.,~\eqref{eq:average}) and that is used to then calculate the log loss ($\log Q_{Y|L}$). On the other hand, for $\cent{P_{Y|\lastlayer}}{Q_{Y|\lastlayer}}$, we evaluate log loss for each sample of $\lastlayer$ separately (i.e., $ \log Q_{Y|\lastlayer}$). These log losses are then averaged to calculate the loss corresponding to each tuple $(x,\ell,y)$.

\section*{Acknowledgements}
The authors gratefully acknowledge discussions with Artemy Kolchinsky and Brendan Tracey, Santa Fe Institute. We are particularly indebted to Christian Knoll, Graz University of Technology, for his help in making our analysis of representational robustness more clear.
This work was supported by the German Federal Ministry of Education and Research in the framework of the Alexander von
Humboldt-Professorship. The work of Bernhard C. Geiger has been funded by the Erwin Schr\"odinger Fellowship J 3765 of the Austrian Science Fund.  The Know-Center is funded within the Austrian COMET Program - Competence Centers for Excellent Technologies - under the auspices of the Austrian Federal Ministry of Transport, Innovation and Technology, the Austrian Federal Ministry of Digital and Economic Affairs, and by the State of Styria. COMET is managed by the Austrian Research Promotion Agency FFG.

\bibliography{references}
\bibliographystyle{IEEEtran}

\begin{IEEEbiography}[{\includegraphics[width=1in,height=1.25in,clip,keepaspectratio]{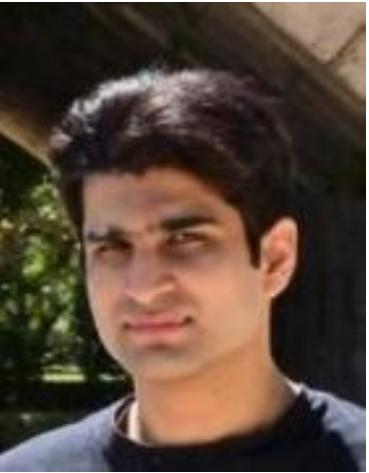}}]{Rana Ali Amjad}
	(S'13) was born in Sahiwal, Pakistan, in 1989. He received the Bachelors degree in Electrical Engineering  (with highest distinction) from University of Engineering and Technology, Lahore, Pakistan, in 2011. He completed his Masters degree in Communication Engineering (with highest distinction) from Technical University of Munich, Germany, in 2013.

	Since 2014 he is pursuing his PhD at the Institute for Communication Engineering at Technical University of Munich. He has received various awards in his academic career including the faculty award for best Master thesis, award for outstanding performance in Master's degree and Gold medal for best performance in Communications major during his Bachelors degree. His research interests cover information theory, machine learning, communication theory, channel coding and information-theoretic security.
\end{IEEEbiography}

\begin{IEEEbiography}[{\includegraphics[width=1in,height=1.25in,clip,keepaspectratio]{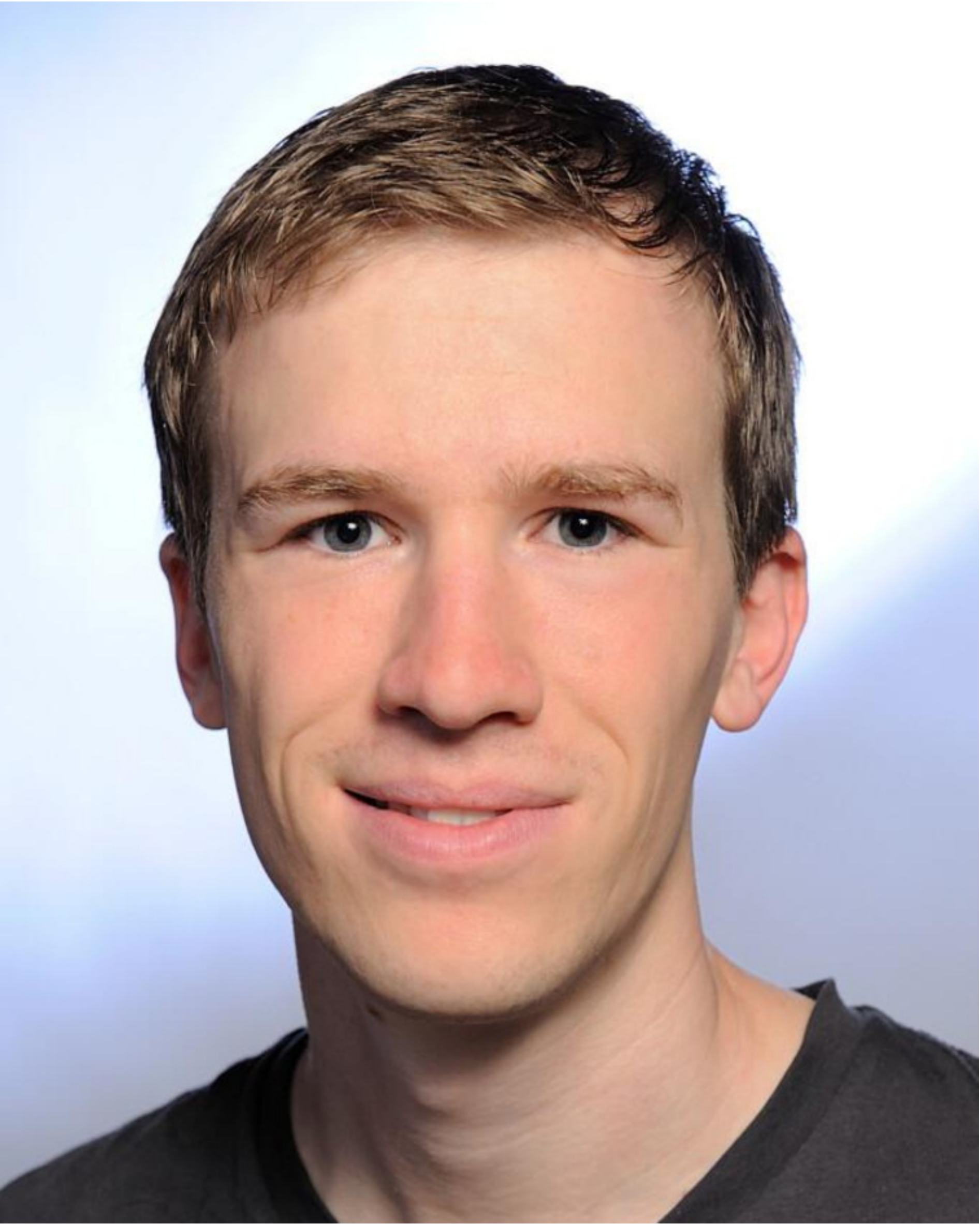}}]{Bernhard C. Geiger}
(S'07, M'14, SM'19) was born in Graz, Austria, in 1984. He received the Dipl.-Ing. degree in electrical engineering (with distinction) and the Dr. techn. degree in electrical and information engineering (with distinction) from Graz University of Technology, Austria, in 2009 and 2014, respectively.

In 2009 he joined the Signal Processing and Speech Communication Laboratory, Graz University of Technology, as a Project Assistant and took a position as a Research and Teaching Associate at the same lab in 2010. He was a Senior Scientist and Erwin Schr\"odinger Fellow at the Institute for Communications Engineering, Technical University of Munich, Germany from 2014 to 2017 and a postdoctoral researcher at the Signal Processing and Speech Communication Laboratory, Graz University of Technology, Austria from 2017 to 2018. He is currently a Senior Researcher at Know-Center GmbH, Graz, Austria. His research interests cover information theory for machine learning and information-theoretic model reduction for Markov chains and hidden Markov models.
\end{IEEEbiography}

\end{document}